\documentclass{article}

    \PassOptionsToPackage{numbers, compress}{natbib}

\usepackage[preprint]{neurips_2023}

\usepackage[utf8]{inputenc} 
\usepackage[T1]{fontenc}    
\usepackage{hyperref}       
\usepackage{url}            
\usepackage{booktabs}       
\usepackage{amsfonts}       
\usepackage{nicefrac}       
\usepackage{microtype}      
\usepackage{xcolor,wrapfig}         
\newcommand{\sv}[1]{\textcolor{black}{#1}}
\newcommand{\updates}[1]{\textcolor{black}{#1}}
\newcommand{\newapp}[1]{\textcolor{black}{#1}}
\usepackage{soul}
\usepackage{url}
\usepackage{epstopdf}
\usepackage{float}
\usepackage[utf8]{inputenc}
\usepackage[pdftex]{graphicx}
\usepackage{multirow,multicol}
\usepackage{amsmath}
\usepackage{amssymb}
\usepackage{caption}
\usepackage{pifont}         
\usepackage{booktabs,bbm,dsfont}
\usepackage{algorithm}
\usepackage{enumitem}
\usepackage{booktabs} 
\usepackage{array}
\usepackage{color,soul,xspace,xcolor}
\usepackage[noend]{algpseudocode}
\algblockdefx[Foreach]{Foreach}{EndForeach}[1]{\textbf{for each} #1 \textbf{do}}{\textbf{end for}}

\algnewcommand{\IIf}[1]{\State\algorithmicif\ #1\ \algorithmicthen}
\algnewcommand{\EndIIf}{\unskip\ \algorithmicend\ \algorithmicif}
\usepackage{comment}
\usepackage{epsfig}
\usepackage{subfig}
\usepackage{mathrsfs,mdwlist,enumitem}

\newcommand{\cmark}{\ding{51}}
\newcommand{\xmark}{\ding{55}}
\newcommand{\cff}{\textsc{Cf}$^2$\xspace}
\newcommand{\cfg}{\textsc{CF-GnnExplainer}\xspace}
\newcommand{\cfx}{\textsc{CF-GnnEx}\xspace}
\newcommand{\name}{\textsc{InduCE}\xspace}
\newcommand{\pgexp}{\textsc{PGExplainer}\xspace}
\newcommand{\gem}{\textsc{Gem}\xspace}
\newcommand{\gnns}{\textsc{Gnn}s\xspace}
\newcommand{\gnn}{\textsc{Gnn}\xspace}
\newcommand{\gat}{\textsc{Gat}\xspace}
\newcommand{\MLP}{\texttt{MLP}\xspace}
\newcommand{\softmax}{\texttt{SoftMax}\xspace}
\newcommand{\CG}{\mathcal{G}\xspace}

\newcommand{\CN}{\mathcal{N}\xspace}
\newcommand{\CT}{\mathcal{T}\xspace}

\newcommand{\CP}{\mathcal{P}\xspace}
\newcommand{\CV}{\mathcal{V}\xspace}
\newcommand{\CE}{\mathcal{E}\xspace}
\newcommand{\CC}{\mathcal{C}\xspace}
\newcommand{\CX}{\boldsymbol{\mathcal{X}\xspace}}

\newcommand{\CS}{\mathcal{S}\xspace}

\newcommand{\CA}{\boldsymbol{\mathcal{A}}\xspace}
\newcommand{\cW}{\mathbf{W}\xspace}
\newcommand{\cS}{\mathbf{S}\xspace}

\newcommand{\cx}{\mathbf{x}\xspace}
\newcommand{\ca}{\mathbf{a}\xspace}

\newcommand{\ch}{\mathbf{h}\xspace}

\setlist{nolistsep,leftmargin=*}

\newtheorem{thm}{\textbf{Theorem}}

\newtheorem{defn}{\textbf{Definition}}

\newtheorem{prob}{\textbf{Problem}}

\title{Empowering Counterfactual Reasoning over Graph Neural Networks through Inductivity}
\author{
Samidha Verma \\
  Indian Institute of Technology, Delhi, India \\
  \texttt{samidha.verma@cse.iitd.ac.in} \\
  \And
  Burouj Armgaan \\
  Indian Institute of Technology, Delhi, India \\
  \texttt{Burouj.Armgaan@cse.iitd.ac.in} \\
  \And
  Sourav Medya \\
  University of Illinois, Chicago, USA\\
  \texttt{medya@uic.edu} \\
  \And
  Sayan Ranu \\
  Indian Institute of Technology, Delhi, India \\
  \texttt{sayanranu@iitd.ac.in} \\
}

\begin{document}

\maketitle
\begin{abstract}

Graph neural networks (\gnns) have various practical applications, such as drug discovery, recommendation engines, and chip design. However, GNNs lack transparency as they cannot provide understandable explanations for their predictions. To address this issue, counterfactual reasoning is used. The main goal is to make minimal changes to the input graph of a \gnn in order to alter its prediction. While several algorithms have been proposed for counterfactual explanations of \gnns, most of them have two main drawbacks. Firstly, they only consider edge deletions as perturbations. Secondly, the counterfactual explanation models are transductive, meaning they do not generalize to unseen data. In this study, we introduce an inductive algorithm called \name, which overcomes these limitations. By conducting extensive experiments on several datasets, we demonstrate that incorporating edge additions leads to better counterfactual results compared to the existing methods. Moreover, the inductive modeling approach allows \name to directly predict counterfactual perturbations without requiring instance-specific training. This results in significant computational speed improvements compared to baseline methods and enables scalable counterfactual analysis for \gnns.

\end{abstract}
\vspace{-0.20in}
\section{Introduction and Related Work}
\label{sec:intro}
\vspace{-0.10in}
The applications of Graph Neural Networks (\gnns) have percolated beyond the academic community. \gnns have been used for drug discovery~\cite{drugdiscovery}, designing chips~\cite{chip}, and recommendation engines~\cite{pinsage}. 
Despite significant success in prediction accuracy, \gnns, like other deep learning based models, lack the ability to explain why a particular prediction was made. Explainability of a prediction model is important towards making it trust-worthy. In addition, it sheds light on potential flaws and generates insights on how to further refine a model. 

\noindent
\textbf{Existing Works: } \sv{At a high level, \gnn explainers can be classified into the two groups of \textit{instance-level}~\cite{gnnexplainer,pgexplainer,rgexplainer,subgraphx,huang2022graphlime,yuan2022explainability, Cfgnnexplainer,cff, gem, rcexplainer, braincf, moleculecf} or \textit{model-level} explanations~\cite{yuan2020xgnn}. 
 Consistent with their nomenclature, instance-level explainers explain a specific input graph, whereas model-level explainers provide a high-level explanation in understanding general behaviour of the \gnn model trained over a set of graphs. 
 Recent research has also focused on global concept-based ~\cite{xuanyuan2023global, azzolin2023global} explainers that provide both model and instance-level explanations.} 
\sv{
Instance-level methods can broadly be grouped into two categories:} \textit{factual} reasoning ~\cite{gnnexplainer,pgexplainer,rgexplainer,subgraphx,huang2022graphlime,yuan2022explainability} and \textit{counterfactual} reasoning~\cite{Cfgnnexplainer,cff, rcexplainer, braincf, moleculecf}. Given the input graph and a \gnn, factual reasoners seek to identify the smallest sub-graph that is sufficient to make the same prediction as on the entire input graph. Counterfactual reasoners, on the other hand, seek to identify the smallest perturbation on the input data that changes the \gnn's prediction. Perturbations correspond to removal and addition of edges. 

Compared to factual reasoning, counterfactual reasoners have the additional advantage of providing a means for recourse \cite{voigt2017eu}. For example, in drug discovery \cite{jiang2020drug, xiong2021graph}, mutagenicity is an adverse property of a molecule that hampers its
potential to become a drug \cite{kazius2005derivation}. While factual explainers can attribute the subgraph causing mutagenecity, counterfactual reasoners can identify this subgraph along with the changes that would make the molecule non-mutagenic. 
 \looseness=-1
\begin{figure}
    \centering
    \includegraphics[width=2.7in]{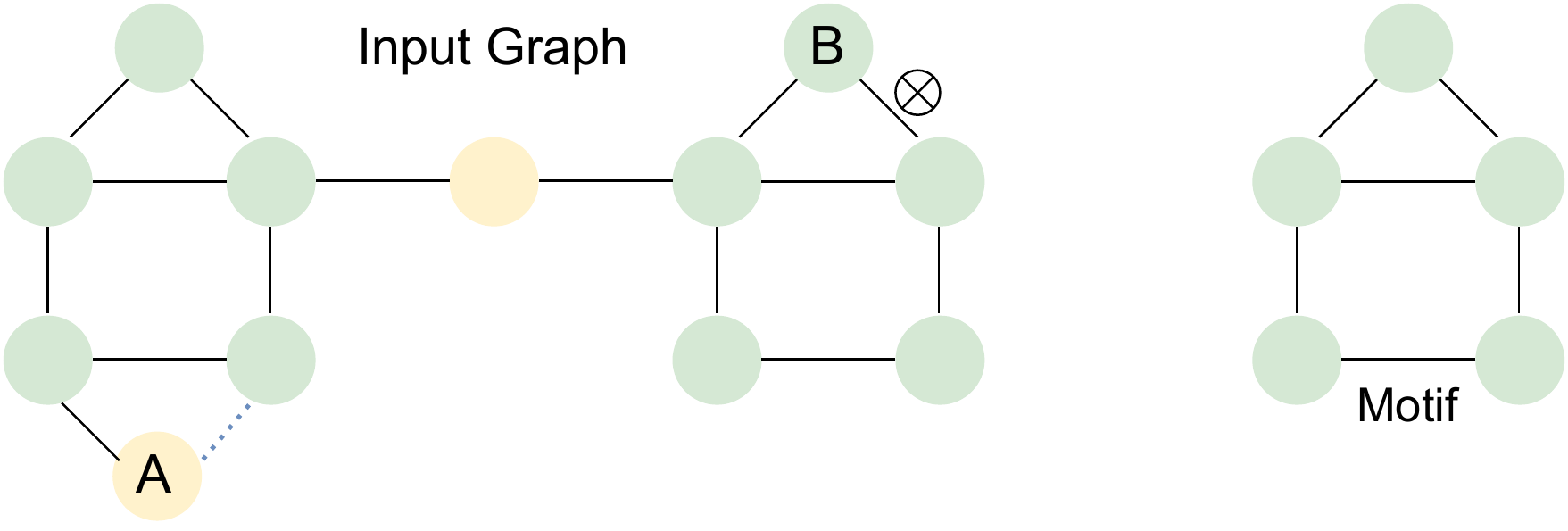}
    \caption{The figure contains two graphs with the right graph being labeled “Motif”. Each node in the left graph belongs to either the green class (label) or yellow. Green class indicates a node that is part of a subgraph isomorphic to the motif; yellow otherwise. Addition of the dotted edge incident on node A changes its label from yellow to green since it becomes part of the motif.}
    \label{fig:motivation}
    \vspace{-0.20in}
\end{figure}

In this work, we study counterfactual reasoning over \gnns towards node classification. 
To illustrate our problem, let us consider the input graph shown in Fig.~\ref{fig:motivation}. Here, each node belongs to the \textit{green} class if it is part of the motif (subgraph) shown on the right. Otherwise, it belongs to the \textit{yellow} class. The dotted edge on node \texttt{A} does not exist, for now. At this stage, if we ask the counterfactual reasoner to flip the label of node \texttt{A}, the best answer would be to add the dotted edge. Similarly, for node \texttt{B}, one possible answer would be to delete the edge marked with $\otimes$. 

Existing works on counter-factual reasoning over \gnns suffer from two key limitations:
\vspace{-0.05in}
\begin{itemize}
    \item {\bf Ability to add edges:} Most of the existing techniques do not consider addition of edges (or nodes); they only consider edge removals. This limitation severely compromises the search space consisting of possible ``changes'' on the input graph. As an example, in Fig.~\ref{fig:motivation}, if we only consider deletions, it is impossible to flip the label of \texttt{A}.
    \item {\bf Inductive modeling:} Existing techniques, with the exception of \gem~\cite{gem}, are transductive in nature, i.e., they cannot generate counterfactuals on unseen nodes. As an example, if the model is trained to generate counterfactuals on node $v$ of graph $G$, it cannot be used to generate counterfactuals on another node $u$ of $G$. Consequently, these transductive models need to be retrained on each node of an input graph. In contrast, an inductive model learns  parameters from a train set of nodes, which in turn can be used to \textit{predict} counterfactual on unseen nodes. 
    In addition, an inductive model is robust to changes in the input graph due to external factors such as new friend connections in a social network, citations in a citation network, etc.
\end{itemize}
\sv{Table~\ref{table:baseline_comparison} in the Appendix presents a structured summary of the instance-level explainers.}

\noindent
\textbf{Contributions: }
In this work, we develop \name (\underline{Indu}ctive \underline{C}ounter-factual \underline{E}xplanations), that addresses the above limitations of existing counterfactual reasoners. 
We propose \name to addresses these  challenges and make the following contributions:
\looseness=-1
\begin{itemize}
    \item \textbf{Novel formulation:} We formulate the novel problem of \textit{model-agnostic, inductive} counterfactual reasoning over \gnns for node classification. It is worth noting that both inductive modeling and the ability to add edges introduce non-trivial challenges. In inductive modeling, we need to learn parameters that embodies general rules to be used for predicting counterfactuals. In the transductive approach, since parameters are learned for each specific node, there is no generalization component. Edge additions introduce a significant scalability challenge as the number of possible additions grows quadratically to the number of nodes in the graph. In contrast, the number of edge deletions is $O(|\CE|)$, where $\CE$ is the set of edges in the graph. (\S~\ref{sec:formulation}).
    \item {\bf Algorithm:} Identifying the smallest number of edge additions or removals that alter the prediction is a combinatorial optimization problem. We prove that 
 computing the optimal solution to the problem is NP-hard~(\S~\ref{sec:formulation}. As a heuristic, we \textit{learn} to solve this combinatorial optimization problem through reinforcement learning powered by \textit{policy gradients}~\cite{reinforce}~(\S~\ref{sec:induce}).
    \item {\bf Empirical validation:} Through extensive experiments on benchmark graph datasets, we show that \name outperforms state-of-the-art algorithms in metrics relevant to counterfactual reasoning. We further analyze the generated counterfactuals and provide compelling evidence that enabling edge additions is indeed the reason driving \name's superior performance. Finally, we also showcase the computation gains obtained due to embracing the inductive paradigm instead of transductive modeling (\S~\ref{sec:experiments}).
\end{itemize}

\section{Preliminaries and Problem Formulation}
\label{sec:formulation}
We use the notation $\CG=(\CV,\CE)$ to denote a graph with node set $\CV$ and edge set $\CE$. We assume each node $v_i\in\CV$ is characterized by a feature vector $x_i\in\mathbb{R}^d$. Furthermore, $l(v):v\rightarrow \CC$ is a function mapping each node $v$ to its true class label drawn from a set $\CC$.
We assume there exists a \gnn $\Phi$ that has been trained on $\CG$. Given an input node $v_i\in \CV$, we assume $\Phi(\CG,v,c)$ outputs a probability distribution over class labels $c\in\CC$. The predicted class label is therefore the class with the highest probability, which we denote as $L_{\Phi}(\CG,v)=\arg\max_{c\in\CC}\{\Phi(\CG,v,c)\}$. 
\looseness=-1
\vspace{-0.05in}
\begin{prob}[Counterfactual Reasoning on \gnns]
\label{prb:cf}
   Given input graph $\CG=(\CV,\CE)$, a target node $v\in\CV$, a \gnn model $\Phi$, and an optional set of node pairs $\mathcal{V}_c=\{(v_i,v_j)\mid v_i,v_j\in\CV\}$ between which edges may be perturbed, find the closest graph $\CG^*$ by minimizing the number of perturbations, such that $L_{\Phi}(\CG^*,v)\neq L_{\Phi}(\CG,v)$ and all perturbed edges are among pairs in $\mathcal{V}_c$.
   \end{prob}

In a real world, we may not have control over all perturbations. $\mathcal{V}_c$ allows us to specify that. If $\mathcal{V}_c\subseteq\CE$, we restrict  to only deletions. On the other hand, if $\mathcal{V}_c\cap\CE=\emptyset$, we only allow additions.

In our problem, we enforce two restrictions on the counterfactual reasoner. First, it should be \textit{model-agnostic}, i.e., only the output of $\Phi$ is visible to us, but not its parameters. Second, the reasoner should be \textit{inductive}, which means we should learn a \textit{predictive model} $\Pi$, that can predict the counterfactual graph $\CG^*$ given the inputs $\CG$, \gnn $\Phi$, and target node $v$. 
\looseness=-1
\vspace{-0.05in}
\begin{thm}[NP-hardness]
\label{thm:nphard}
Counterfactual reasoning for \gnns, i.e., Prob.~\ref{prb:cf}, is NP-hard.
\end{thm}

We prove NP-hardness by mapping counter-factual reasoning over \gnns to the \textit{set-cover} problem. The details The are provided in the App.~\ref{sec:app_proof_nphard}). Owing to NP-hardness, it is not feasible to identify the closest counterfactual graph in polynomial time. Hence, we aim to design effective heuristics.

 \vspace{-0.15in}
\section{\name}
\label{sec:induce}
\vspace{-0.15in}
Our goal is to learn an inductive counterfactual reasoning model $\Pi$, and thus, the proposed algorithm is broken into two phases: \textit{training} and \textit{inference}. During training, we learn the parameters of the model $\Pi$ and during inference, we predict the counterfactual graph using $\Pi$. 
Theorem~\ref{thm:nphard} prohibits us from supervised learning since generating training data of ground-truth counterfactuals is NP-hard. Hence, we use reinforcement learning. Through \textit{discounted rewards}, reinforcement learning allows us to model the combinatorial relationships \cite{khalil2017learning} in the perturbation space. 
\vspace{-0.10in}
\subsection{Learning $\Pi$ as an MDP}
\label{sec:training}
\vspace{-0.05in}
Given graph $\CG$, we randomly select a subset of vertices from $\CV$ to train $\Pi$. Given a target node $v$, the task of $\Pi$ is to iteratively delete or add edges such that with each perturbation the likelihood of $\Phi(\CG^t,v)\neq \Phi(\CG,v)$ changes maximally. Here, $\CG^t=(\CV,\CE^t)$ denotes graph $\CG$ after $t$ perturbations starting with $\CG^0=\CG$. We model this task of iterative perturbations as a \textit{markov decision process (MDP)}. Specifically, the \textit{state} captures a latent representation of the graph indicative of how it would react to a perturbation. An \textit{action} corresponds to an edge addition or deletion by $\Pi$. Finally, the \textit{reward} is a function of the number of perturbations, which we want to minimize, and the probability of 
 $\Phi(\CG^t,v)$ flipping following the next action (edge addition or deletion), a value that we want to maximize. We next formalize each of these notions.
 \looseness=-1

 

\noindent
\textbf{State:} Intuitively, the state should characterize how likely the class label of the target node $v$ would flip following a given action. Towards that end, we observe that a \gnn of $\ell$ layers aggregates information from the $\ell$-hop neighborhood of $v$. Nodes outside this neighborhood do not impact the prediction of a \gnn.
Motivated by this design of \gnns, the state in our problem is the set of node representations in the $h$-hop neighborhood of the target node $v$, where ideally $h>\ell$. Specifically, at time $t$, the state is:
\vspace{-0.15in}
\begin{alignat}{2}
\cS^t_v&=\left\{\cx^t_u \mid u\in\CN^{h}_v\right\}\text{, where}\\
\CN^{h}_v&=\left\{u\in\CV\mid sp(v,u)\leq h\right\}
\end{alignat}
Here, $sp(v,u)$ denotes the length of the shortest path from $v$ to $u$ in the original graph $\CG=\left(\CV,\CE\right)$. 

The representations of nodes, i.e., $\cx^t_u$, are constructed using a combination of semantic, topological, and statistical features. 
\looseness=-1
\begin{figure*}
\vspace{-0.4in}
    \centering
\includegraphics[width=5.8in]{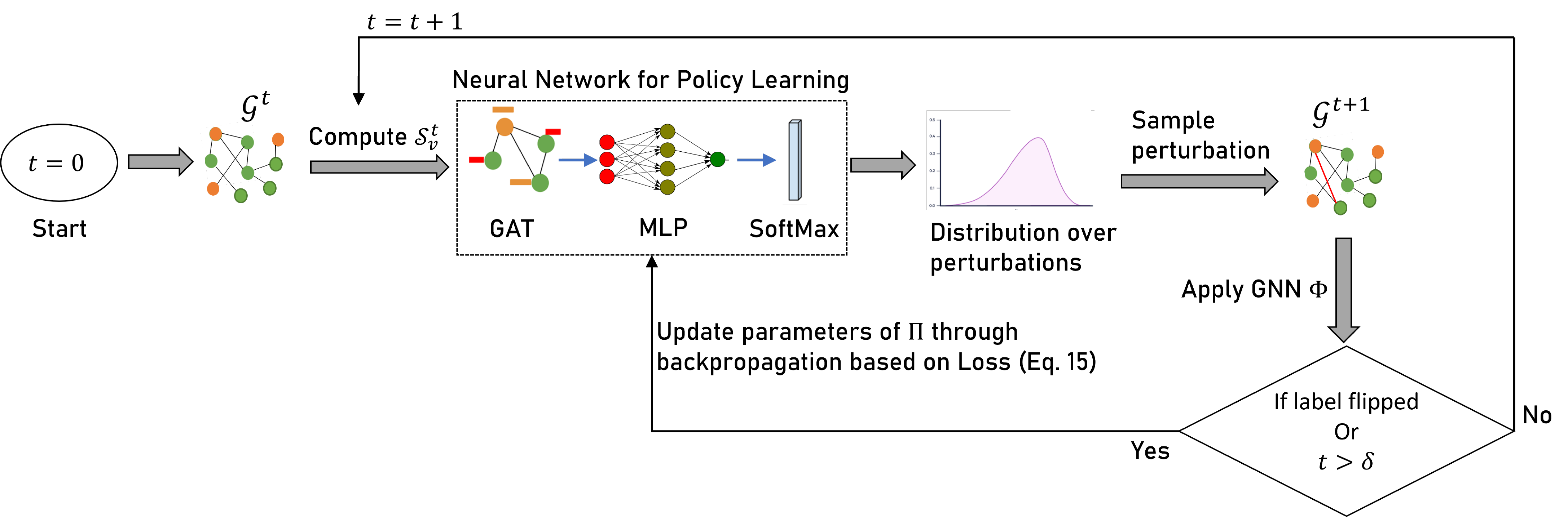}
    \vspace{-0.2in}
    \caption{Pipeline of the policy learning algorithm in \name. $\delta$ indicates the maximum number of allowed perturbations.}
    \label{fig:induce}
    \vspace{-0.20in}
\end{figure*}
\begin{itemize}
\item \textit{Original node Features: } It is common to encounter graphs where nodes are annotated with features or labels (recall the definition of $x_i$ in \S~\ref{sec:formulation}). We retain these features. 
\item \textit{Degree Centrality: }The higher the degree of a node, the more information it receives from its neighbors. Thus, when an edge is added or deleted from the target node to a high-degree node, it may have significant impact on the representation of the target. Based on this observation, we use the degree of a node as a part its representation.
\looseness=-1
\item \textit{Entropy: }The entropy of a node at time $t$ is defined as $e^t_u = -\sum_{\forall c\in \CC} p_c\log p_c\text{, where }
    p_c = \Phi \left(\CG^t,v,c\right)$. The entropy quantifies the uncertainty of the \gnn $\Phi$ on a given node. We hypothesize that if the $\Phi$ is highly certain (i.e., low entropy) about the class label of some node $u$, then any perturbation on $u$ is unlikely to make it flip. Similarly, the opposite is true on nodes with high entropy. Due to this information content of entropy, we use it as one of the features in $\cx^t_u$.


\item \textit{Class label: } Finally, we include the predicted class label of a node, i.e., $L_{\Phi}(\CG^t,u)$ in the form of one-hot encodings of dimension $\CC$.

\noindent
\end{itemize}

\noindent
The final representation of node $u$ and time $t$ is therefore the concatenation of the above features, i.e.,
\updates{
\begin{equation}
\label{eq:rep}
    \cx_u^t = x_i \mathbin\Vert degree^t_u \mathbin\Vert e^t_u \mathbin\Vert \left(\text{one-hot}\left(L_{\Phi}\left(\CG^t,u\right)\right)\right)
\end{equation}
Here, $\mathbin\Vert$ represents the \textit{concatenation} operator.} 

\textbf{Actions:} The action space consists of all possible edge deletions in the $h$-hop neighborhood of target node $v$ and additions of edges from $v$ to other non-attached nodes in its $h$-hop neighborhood. Formally, the sets are defined as follows:
\begin{alignat}{2}
\label{eq:del}
\CE^t_{v,del}&=\left\{e=(u_i,u_j)\in\CE^t\mid u_i,u_j\in\CN^{h}_v\right\}\\
\label{eq:add}
\CE^t_{v,add}&=\left\{e=(v,u_j)\not\in\CE^t\mid u_j\in\CN^{h}_v\right\}
\end{alignat}
The action space is the perturbation set:
\vspace{-0.20in}
\begin{equation}
\label{eq:perturbations}
    \hspace{0.09in} \CP^t = \CE^t_{v,del} \cup \CE^t_{v,add}
\end{equation} 
\textbf{Reward:} Our objective
is to flip the predicted label of target node $v$ with the minimum number of perturbations in $\CN^{h}_v$ in order to find the counterfactual. To capture these intricacies, we formulate the reward of an action $a$ as a combination of the prediction accuracy of \gnn $\Phi$ and the number of perturbations made so far. 
\vspace{-0.10in}
\sv{
\begin{alignat}{3}
    \mathcal{R}^t_v(a) &= \frac{1}{\mathcal{L}^{t+1}_{v,{pred}} + \beta\times d(\CG, \CG^t)}\text{, where} \\
      \mathcal{L}^t_{v,pred} &= \sum_{\forall c\in\CC}\mathds{1}_{l(v)=c}\log\left(\Phi\left(\CG^t,v,c\right)\right), d(\CG, \CG^t) &= t+1
\end{alignat} 
}
In simple terms, $\mathcal{L}^{t+1}_{v,pred}$ is the \textit{log-likelihood} of the data predicted by $\Phi$ in $\CG^{t+1}$ on $v$. $\CG^{t+1}$ is the created upon perturbing $\CG^t$ with action $a$. 
$\beta$ is a hyper-parameter that regulates how much weight is given to log-likelihood of the data vs. the  perturbation count. \sv{$d$ is the distance function, which in our case is simply the number of edge edits made to $\CG$ at time step $t$.} 

\noindent
\textbf{State Transitions:} At time $t$, the action corresponds to selecting a perturbation $a\in \CP^t$ (Recall Eq.~\ref{eq:perturbations}) from $p^t_{a,v}\sim\Pi\left(a\mid \CS^t_v\right)$. We will discuss the computation of $p^t_{a,v}$ in \S~\ref{sec:selection}.
\vspace{-0.15in}
\subsection{Neural Architecture for Policy Training}
\label{sec:selection}
To learn $p^t_{a,v}$, we we take the representations in $\cS^t_v$, and pass them through a neural network comprising of a $K$-layered \textit{Graph Attention Network }(\gat)~\citep{gat}, an \MLP, and a final \softmax layer. The \gat learns a $d$-dimensional representation $\ca\in\mathbb{R}^{d}$ for each perturbation $a\in\CP^t_{v}$. $\ca$ is then passed through an \textit{Multi-layered Perceptron (\MLP)} to embed them into a scalar representing their value, which is finally passed over a \softmax layer to learn a distribution over $\CP^t_v$. The entire network is trained end-to-end. We next detail each of these components.

\noindent
\textbf{\gat:} Let $\forall u\in\CN^{h}_v, \ch^0_u=\cx^t_u$ (Recall Eq.\ref{eq:rep}). In each layer $k\in[1,K]$, we perform the following transformation:
\begin{equation}
    \ch^k_u=\sigma\left(\sum_{\forall u'\in \CN^{1}_u\cup\{u\}}\alpha^k _{u,u'}\cW^k\ch^{k-1}_{u'}\right)
\end{equation}
$\sigma$ is an activation function, $\alpha^k _{u,u'}$ are learnable, layer-specific attention weights, and $\cW^k\in\mathbb{R}^{d^{k-1}\times d^k}$ is a learnable, layer-specific weight matrix where $d^k$ is the hyper-parameter denoting the representation dimension in hidden layer $k$. In our implementation, we use LeakyReLU with negative slope $0.01$ as the activation function. The attention weights are learned through an \MLP followed by a \softmax layer. Specifically,
\vspace{-0.10in}
\begin{align*}
e^k_{u,u'}=\MLP(\ch^{k-1}_u \mathbin\Vert  \ch^{k-1}_{u'})\text{, where }e^k_{u,u'}\in\mathbb{R}, \hspace{0.7in}
\alpha^k_{u,u'}=\frac{exp\left(e^k_{u,u'}\right)}{\sum_{\hat{u}\in\CN^{1}_u\cup\{u\}}exp\left(e^k_{u,\hat{u}}\right)}
\end{align*}

 After $K$ layers, the \gat outputs the final representation $\CX_u=\ch^K_u$ for each node $u$ in $v$'s neighborhood. Semantically, given the initial state representation $\cx^t_u$, the \gat enriches them further by merging with topological information. Finally, the representation of an action $a\in \CP^t_v$ is set to $\ca=\mathcal{X}_u\ || \mathcal{X}_v\ ||\ t(u,v)$, where:
\vspace{-0.10in}
\begin{equation}
        t(u,v) = 
    \begin{cases}
      0, &  a \in \CE^t_{v,del}  \text{ (Recall Eq.~\ref{eq:del})}\\
      1, & a \in \CE^t_{v,add} \text{ (Recall Eq.~\ref{eq:add})}
    \end{cases}
\end{equation} 
 
 \noindent
 \textbf{MLP and SoftMax layers:} The value of $a$ is $s_a=\MLP(\ca)$, where $s_a\in\mathbb{R}$. Finally, we get a distribution over all actions in $\CP^t_v$ as:
\vspace{-0.05in}
\begin{equation}
\label{eq:action}
p^t_{a,v}=\Pi(a\mid \CS^t_v)=\frac{exp(s_a)}{\sum_{\forall a'\in \CP^t_v}exp(s_{a'})}
\end{equation}
\vspace{-0.05in}

\vspace{-0.25in}
\subsection{Policy Loss Computation} 
\label{sec:loss}
\vspace{-0.1in}
We iteratively sample an action as per Eq.~\ref{eq:action} till either the label flips or we exceed the maximum number of perturbations (which is a hyper-parameter). This iterative selection generates a trajectory of perturbations $\CT_v=\{a_1,\cdots,a_m\}$. 
 We use the standard loss for policy gradients on $\CT_v$~\cite{reinforce}. More specifically, we minimize the following loss function:
\begin{equation}
\label{eq:loss}
\mathcal{J}(\Pi) =-\frac{1}{\CV_{tr}}\left(\sum_{\forall v\in\CV_{tr}}\left(\sum_{t=0}^{\lvert \CT_v \rvert}\log{p^t_{a,v}} \mathcal{R}^t_v(a_t)+\eta Ent(\CP^t_v)\right)\right)
\end{equation}
Here, $\CV_{tr}\subseteq\CV$ is the subset of nodes on which the RL policy is being trained. $Ent(\CP^t_v)$ is the entropy of the current probability distribution over the action space.
\begin{equation}
Ent(\CP^t_v)=-\sum_{\forall a\in\CP^t_v}p^t_{a,v}\log(p^t_{a,v})
\end{equation}
By adding the entropy to the loss, we encourage the RL agent to explore when there is high uncertainty. $\eta$ is a hyper-parameter balancing the \textit{explore-exploit} trade-off. For simplicity of exposition, we omit the discussion on discounted rewards in Eq.~\ref{eq:loss}. Discounted rewards better capture the combinatorial relationship in the perturbation space. Refer to App.~\ref{app:loss} for details.
\subsection{Training and Inference}
\label{sec:training_inf}
Fig.~\ref{fig:induce} presents the  training pipeline. Starting from the original graph, we compute the state representation at each iteration $t$. The state is passed to the neural network to compute a distribution over the perturbation space. A perturbation is sampled from this distribution and the graph is accordingly modified. The \gnn $\phi$ is then applied on the modified graph. If the label flips or the number of perturbation exceeds the maximum limit, we update the policy parameters. Otherwise, we update the state and continue building the perturbation trajectory in the same manner. The pseudocode of the training pipeline is provided in Alg.\ref{alg:induce} in the Appendix.
\looseness=-1

\noindent
\textbf{Inductive inference:} We iteratively make forward passes till the label flips or we exceed the budget. The forward pass is identical to the training phase with the only exception being we deterministically choose the perturbation with the highest likelihood instead of sampling.

\noindent
\textbf{Transductive Inference:} This phase proceeds identical to the training phase with the only exception that we learn a target node specific policy instead of one that generalizes across all nodes.

\textbf{Complexity of \name}: The time complexity of training phase is  $\mathcal{O}(|\mathcal{V}_{tr}|(\mathcal{|V|} + \mathcal{|E|}))$ and the test phase is $\mathcal{O}(|\mathcal{V}|_{test}|(\mathcal{|V|} + \mathcal{|E|}))$. Here $\CV_{tr}$ and $\CV_{test}$ denote the number of nodes in the train and test sets respectively. The derivations are provided in App.~\ref{sec:app_complexity}. 

\section{Experiments}
\label{sec:experiments}
In this section, we benchmark \name against established baselines. The code base and datasets used in our evaluation are available anonymously at \url{https://github.com/idea-iitd/InduCE.git}. Details of the hardware and software platform are provided in App.~\ref{app:setup}.
\vspace{-0.10in}

\noindent
\subsection{Datasets}
\label{sec:datasets}
\begin{table}
\centering
\caption{The statistics of the benchmark datasets.}
\label{table:dataset}
\scriptsize
\begin{tabular}{ lccccc } 
 \toprule
  & \textbf{Tree-Cycles} & \textbf{Tree-Grid} & \textbf{BA-Shapes} & \textbf{Amazon} & \textbf{ogbn-arxiv}\\
 \midrule
 \# Classes & 2 & 2 & 4 & 6 & 40\\
 \# Nodes & 871 & 1231 & 700 & 397& 169,343\\
 \# Edges & 1950 & 3410 & 4100 & 2700& 1,166,243\\
 Motif size (\# nodes) & 6 & 9 & 5 & NA& NA\\
 Motif size (\# edges)& 6 & 12 & 6 & NA& NA\\
 \# Nodes from motifs &360  &720  & 400 & NA & NA\\
Avg node degree & 2.23 & 2.77 & 5.86 & 15.90 & 6.89\\
\hline
\end{tabular}
\end{table}
\textbf{Benchmark Datasets:} 
We use the same three benchmark graph datasets used in \cite{cff,gem,Cfgnnexplainer}. Statistics of these datasets are listed in Table~\ref{table:dataset}. Each dataset has an undirected base graph with pre-defined motifs attached to random nodes of the base graph,
and randomly added additional edges
to the overall graph. The class label of a node indicates whether it is part of a node or not. Further details on the datasets are provided in App.~\ref{app:datasets}.

\noindent
$\bullet$ \textbf{Real Dataset:} We additionally use \updates{real-world datasets from the Amazon-photos co-purchase network~\cite{shchur2018pitfalls} and ogbn-arxiv~\cite{10.1162/qss_a_00021}}. In the Amazon dataset, each node corresponds to a product, edges correspond to products that are frequently co-purchased, node features encode bag-of-words from  product reviews and the node class label indicates the product category. 
\updates{The ogbn-arxiv dataset is a citation network. The nodes are all computer science arXiv papers indexed by MAG~\cite{10.1162/qss_a_00021}. Each directed edge represents that one paper cites another. The features are word embeddings of the title and the abstract computed by the skip-gram model~\cite{NIPS2013_9aa42b31}. The labels are subject areas. Since the class labels in these datasets are not based on presence or absence of motifs, the corresponding cells in Table~\ref{table:dataset} are marked as ``NA''.}
\vspace{-0.10in}
\subsection{Baselines} 
\vspace{-0.10in}
We benchmark \name against the state-of-the-art baselines of \textbf{(1)} \cfg~\cite{Cfgnnexplainer}, \textbf{(2)} \cff~\cite{cff}, and \textbf{(3)} \gem~\citep{gem}. In addition, we also compare against the state-of-the-art factual explainer \textbf{(4)} \pgexp to show that when factual explainers are used for counter-factual reasoning by removing the factual explanation (subgraph) from the input graph, they are not effective. This is consistent with prior reported literature~\cite{Cfgnnexplainer, cff, gem}. Finally, we also compare against \textbf{(5)} \textsc{Random} perturbations. While \cff and \cfg are  transductive, \gem and \pgexp are inductive. The codebase of all algorithms have been obtained from the respective authors.




\sv{We do not consider \cite{xgnn_kdd20} and \cite{rcexplainer}} since they are limited to graph classification. We omit \textsc{GnnExplainer}\cite{gnnexplainer}, since both \cff and \cfg outperformed \textsc{GnnExplainer}. Further, we do not study Bacciu et al. \cite{bacciuetal} 
since it uses internal representations of the black-box \gnn model to exploit domain-specific knowledge. We focus on a domain-agnostic setting. 
Furthermore, unlike in \cite{bacciuetal}, we do not assume access to the embeddings of the black-box model. \sv{Hence, our algorithm is also applicable in situations where the internal details of the \gnn are hidden from the end-user due to proprietary reasons.}


\subsubsection{Performance measures}
To quantify performance, we use the standard measures from the literature \cite{Cfgnnexplainer}. 
\begin{itemize}
\item \textbf{Fidelity:} Fidelity is the percentage of nodes whose labels do not change when the edges produced by the explainer (algorithm) are perturbed. Lower fidelity is better. Furthermore, it may be argued that fidelity is the most important metric among the three measures.
    \item \textbf{Size:} 
    Explanation size is the number of edges perturbed for a given node. Lower size is better. 
    \item \textbf{Accuracy:} 
    Accuracy is the percentage of explanations that are correct. As standard in \cff,\cfg, and \gem, this translates to the percentage of edges in the counterfactual that belong to the motif. Since nodes have a non-zero class label only if they belong to a motif, the explanation for nodes should be edges in the motif itself. Note that accuracy is computable only on the benchmark datasets since they include ground-truth explanations.
    \sv{ \item \textbf{Sparsity: }Sparsity is defined as the proportion of edges from $\CN^{\ell}_v$, i.e., the $\ell$-hop neighbourhood of the target node, that is retained in the counter-factual $v$~\cite{yuan2022explainability}; a value close to $1$ is desired. Since sparsity is inversely correlated to size, we present sparsity values of our experiments in App.~\ref{app:sparsity}.}

    
\end{itemize}
\vspace{-0.05in}\textbf{Other settings:} Details of additional experimental settings regarding the counterfactual task, the black-box \gnn, training and inference are given in App.~\ref{app:parameters}.

   \begin{table}[t]
\vspace{-0.25in}
\scriptsize
\scalebox{0.87}{
\begin{tabular}{ l||c|c|c||c|c|c||c|c|c } 
\toprule
 \multirow{2}{*}{\textbf{Method}} & \multicolumn{3}{c||}{\textbf{\textbf{Tree-Cycles}}}  & \multicolumn{3}{c||}{\textbf{Tree-Grid}} & \multicolumn{3}{c}{\textbf{BA-Shapes}}\\\cline{2-10} 
& \textbf{Fid.(\%)} $\downarrow$ & \textbf{Size }$\downarrow$ & \textbf{Acc.(\%)} $\uparrow$ & \textbf{Fid.(\%)} $\downarrow$ &  \textbf{Size} $\downarrow$ & \textbf{Acc.(\%)} $\uparrow$ & \textbf{Fid.(\%)} $\downarrow$ &  \textbf{Size} $\downarrow$ & \textbf{Acc.(\%)}$\uparrow$\\
\midrule
\multicolumn{1}{p{1.7cm}||}{ \textbf{\textsc{Random}}} & \textbf{0} & 3.18 \textpm 2.32 & 67.08 & \textbf{0} & 8.32 \textpm 4.95 & 73.44 & \textbf{0} & 283.97 \textpm 272.76 & 15.57\\
\multicolumn{1}{p{1.7cm}||}{ \textbf{\cfx}} & 49.0 & 1.05 \textpm 0.23 & \textbf{100} & 10 &	1.37 \textpm 0.58 & 92.24 & 37.0 & 1.31 \textpm 0.55 & \textbf{95.83}\\ 
\multicolumn{1}{p{1.7cm}||}{ \textbf{\cff}} & 76.38 & 4.18 \textpm 1.89 & 67.68 & 98.45 & 5.5 \textpm 1.5 & 44.64 & 23.68	& 4.10 \textpm 1.64 & 70.54 \\ 
\multicolumn{1}{p{1.7cm}||}{ \textbf{\name (transductive)}} & \textbf{0} & \textbf{1.01} \textpm \textbf{0.12} & 98.61 & \textbf{0} & \textbf{1.02} \textpm \textbf{0.12} & \textbf{97.67} & \textbf{0} & \textbf{1.30} \textpm \textbf{0.90} & 95.31\\
\midrule
\multicolumn{1}{p{1.8cm}||}{\textbf{\cfx$++$}} & 100 & NULL & NULL & 100 &	NULL & NULL & 38.16 & 	6315.44 \textpm 9916.50 & 17.36\\ 
\multicolumn{1}{p{1.7cm}||}{ \textbf{\cff$++$}} & 13.89 & 28.34 \textpm 7.56  & 19.24 & 28.68 & 12.90 \textpm 7.71 & 27.44 & 100	& NULL & NULL \\ 
\multicolumn{1}{p{1.7cm}||}{\textbf{\name (transductive)$--$}}& \textbf{0} & 1.40 \textpm 1.49 & 81.94 & \textbf{0} & 1.24 \textpm 0.43 & 92.64 & 6.6 & 1.42 \textpm 1.49 & 83.22 \\

 \bottomrule
\end{tabular}}
\caption{\textbf{Results for transductive methods:} Lower fidelity, smaller size, and higher accuracy are desired. The best results are highlighted in bold. Fid. denotes fidelity and Acc. denotes Accuracy.}
\label{table:node_class_results_transductive}
\end{table}

\begin{table}[t]
\centering
\scalebox{0.77}{
\footnotesize
\begin{tabular}{ l||c|c|c||c|c|c||c|c|c } 
\toprule
 \multirow{2}{*}{\textbf{Method}} & \multicolumn{3}{c||}{\textbf{\textbf{Tree-Cycles}}}  & \multicolumn{3}{c||}{\textbf{Tree-Grid}} & \multicolumn{3}{c}{\textbf{BA-Shapes}}\\\cline{2-10} 
 & \textbf{Fid.(\%)} $\downarrow$ & \textbf{Size }$\downarrow$ & \textbf{Acc.(\%)} $\uparrow$ & \textbf{Fid.(\%)} $\downarrow$ &  \textbf{Size} $\downarrow$ & \textbf{Acc.(\%)} $\uparrow$ & \textbf{Fid.(\%)} $\downarrow$ &  \textbf{Size} $\downarrow$ & \textbf{Acc.(\%)}$\uparrow$\\
\midrule
 \multicolumn{1}{p{2cm}||}{ \textbf{\pgexp}} & 34.72 & 6 & 76.85 & 41.09 & 6 & 66.93 &  6.58 & 6 & 89.25\\ 	%
 
 \multicolumn{1}{p{2cm}||}{ \textbf{\gem}} & 95 & 6 & 88.97 & 97 & 6 & \textbf{94.57} &  17 & 6 & \textbf{98.44}\\ 	%
\multicolumn{1}{p{2cm}||}{\textbf{\name (inductive)}} & \textbf{0} & \textbf{2.31} \textpm \textbf{1.44} & \textbf{96.65} & \textbf{0} & \textbf{4.67} \textpm 2.91 & 91.05 & \textbf{2.6} & \textbf{4.37} \textpm 3.53 & 64.40 \\  
\midrule
\multicolumn{1}{p{2cm}||}{\textbf{\name (inductive) \textendash{ 
 }\textendash}} & 36.3 & \textbf{1.67 +- 0.90} & 90.32 & 16.3 & 6.38 \textpm 3.74 & 86.31 & 40.8 & \textbf{3.37 \textpm 3.04} & 56.08\\	
\bottomrule
\end{tabular}}
\caption{\textbf{Results for inductive methods.} The best result in each category is highlighted in bold.
}
\label{table:node_class_inductive}
\end{table}
\subsection{Quantitative Results on Benchmark Datasets}
\label{sec:quantitaive_results}

\noindent
\textbf{Transductive methods:} Table \ref{table:node_class_results_transductive} presents the results (for now, we will focus on the first four rows). Our method \name in the transductive setting outperforms all the baselines almost in all settings. For Tree-Cycles and BA-Shapes, \cfg is producing better accuracy. However, we note that its fidelity is much worse, indicating it fails to find an explanation more frequently. More generally, while \cfg consistently achieves the lowest size among the baselines, its fidelity is much worse. This indicates that \cfg is able to solve only the easy cases and hence the low size is deceptive as it did not solve the difficult ones. 

\noindent
\textbf{Inductive methods:} 
Table~\ref{table:node_class_inductive} shows that \name is superior to \gem and \pgexp in most cases. The fidelity scores produced by \gem and \pgexp are much higher (worse). This indicates, in most of the cases, \gem and \pgexp are unable to find a counterfactual example. 
Also recall that the explanation size is fixed in \gem and \pgexp since they work with fixed budgets.
\looseness=-1
\begin{figure}{r}{3.4in}
    \centering
    \subfloat[Inductive]{\includegraphics[width=1.6in]{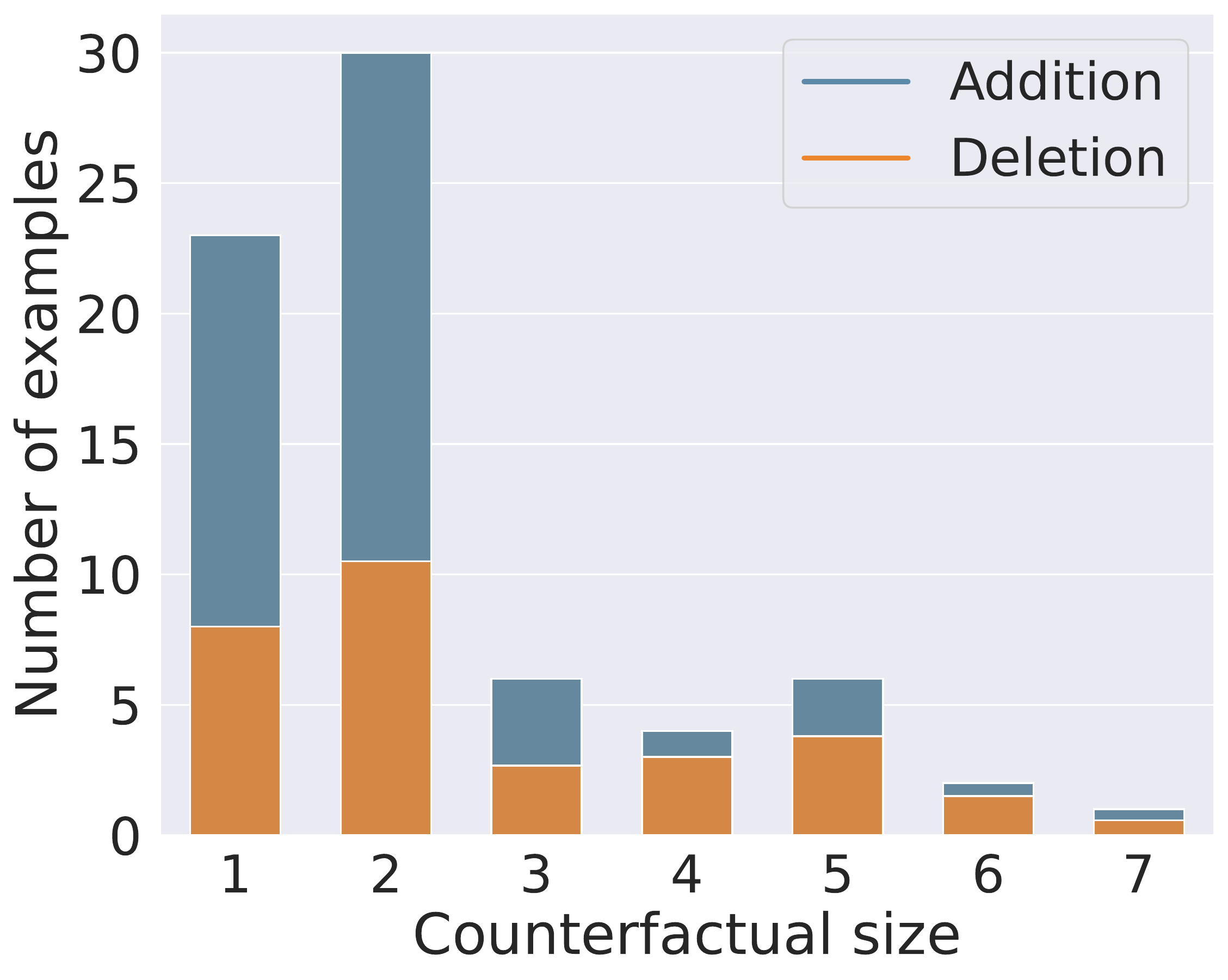}}
    \subfloat[Transductive]{\includegraphics[width=1.8in]{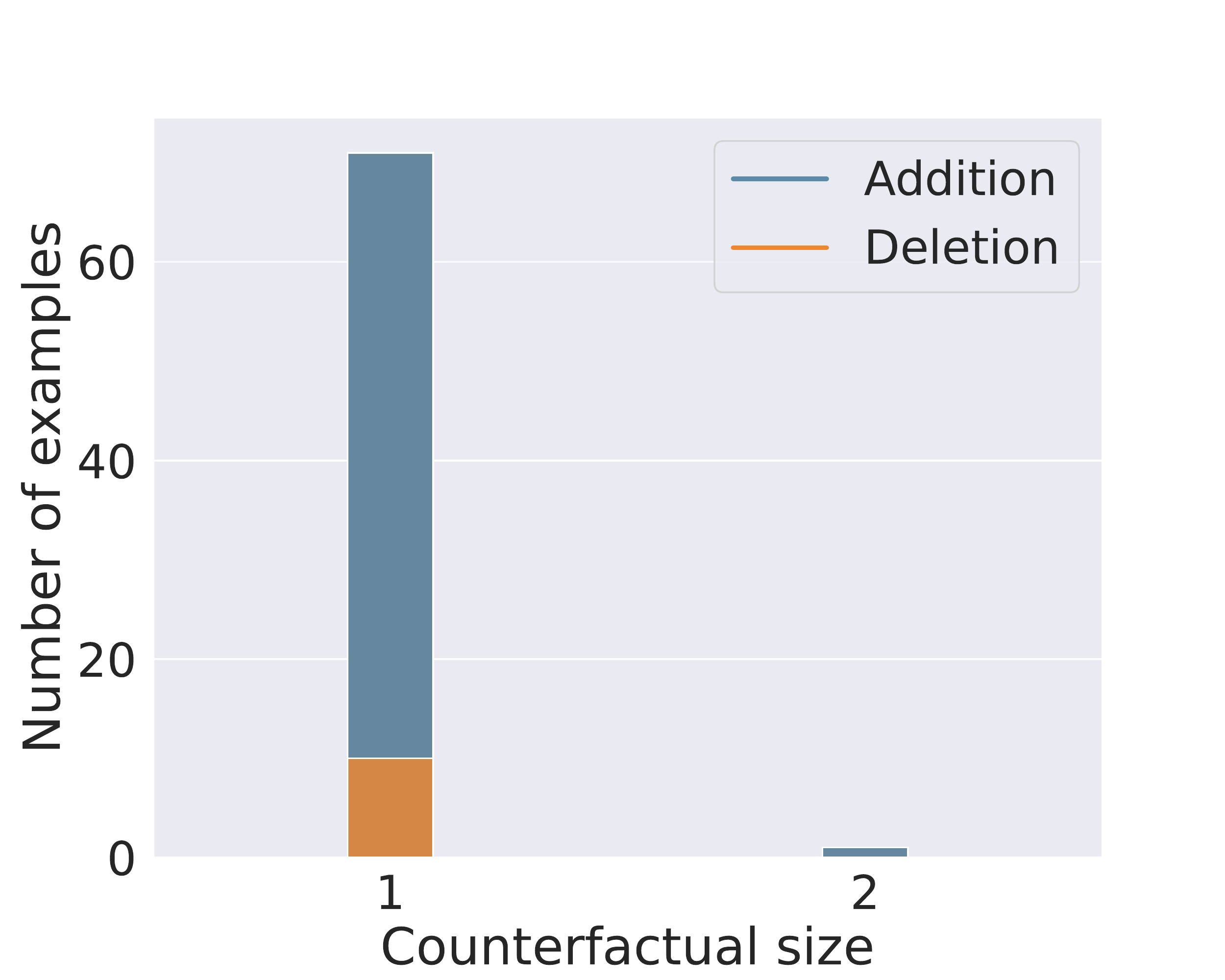}}
    \caption{ The distributions of the edit size and their internal composition of edge additions and deletions by \name on the Tree-Cycles dataset.}
\label{fig:tree_cycle_ind_trans_cf_example}
\end{figure}
\noindent
\textbf{Transductive vs Inductive: } We further compare the inductive version (Table \ref{table:node_class_inductive}) of our method, \name with the transductive baselines (Table  \ref{table:node_class_results_transductive}). While the transductive methods have a clear advantage of re-training the model instance wise, the results produced by \name-inductive are comparable. As noted earlier, although \cfg achieves better size than \name-inductive, its fidelity is much worse indicating that the low size is a manifestation of not being able to explain the hard cases that \name is able to explain. Moreover, in addition to the ability to generalize to unseen nodes, inductive modeling also imparts a dramatic speed-up in generating explanations (see Table \ref{table:running_time}).

\textbf{Impact of edge additions:} We seek answers to two key questions: \textbf{(1)} How much does the performance of \name deteriorate if we restrict edge additions? \textbf{(2)} If we empower the baselines also with additions, do they match up to \name? To answer the first question, we study the performance of \name in the setting where only edge deletions are allowed. The rows corresponding to \name (transductive)$--$ and \name (inductive)$--$ in Tables~\ref{table:node_class_results_transductive} and ~\ref{table:node_class_inductive} present these results. It is evident that the deletion-only version produces inferior results for both the transductive and inductive versions. In Fig.~\ref{fig:tree_cycle_ind_trans_cf_example}, we further study the frequency distribution of edge additions and deletions in the counter-factual explanations produced by \name in Tree-Cycles dataset (results on other datasets are in App. \ref{app:counterfactual_size_disn}). We observe that additions dominate the perturbations, and thereby, further establishing its importance, which \name unleashes.  

To address the second question, we empower \cfg and \cff with edge additions, denoted as \cfg$++$ and \cff$++$ respectively. \footnote{\gem is not extendible to additions (See App.~\ref{app:baselines} for details), \pgexp does not incorporate perturbations with the intent of flipping the label since it is a factual explainer.} Both \cff and \cfg use a \textit{mask-based} strategy. A mask is a learnable binary matrix of the same dimension as the $\ell$-hop neighborhood of the target node. By taking an element-wise product of the mask with the adjacency matrix, one obtains the edges to be deleted. When empowered with additions, the mask itself becomes the new adjacency matrix. Surprisingly, the performance of \cfg drops, while for \cff, we see improvement in fidelity in two out of three datasets. Further investigation into this performance reveals that edge additions significantly increase the search space of possible perturbations (See Table~\ref{tab:actionspace} in Appendix). A mask-based strategy is a single-shot learnable paradigm that does not examine the marginal effect of each perturbation. When the perturbation space increases, it overwhelms the learning procedure. In contrast, \name uses reinforcement learning where a trajectory of perturbations is selected based on their marginal gains. This allows better modeling of the combinatorial nature of counter-factual reasoning.

Overall, the above experiments reveal that both additions, as well as an algorithm equipped to model large combinatorial spaces, are required to perform well.

\textbf{Additional experiments:} App.~\ref{app:ablation} contains further empirical data on \textbf{(1)} the impact of heuristic features and \textbf{(2)} the choice of \gnn architecture in the MDP on performance. \updates{Experiments on counterfactual size vs accuracy trade-off are given in App. \ref{app:size_vs_acc}.}

\vspace{-0.10in}
\subsection{Quantitative Results on Real Datasets}
\label{sec:quantitative_results_amazon}
\vspace{-0.10in}
\begin{table}[t]
\vspace{-0.30in}
\subfloat[Transductive]{
\label{tab:transductive_all_amazon}
\scalebox{0.7}{
\begin{tabular}{ c | c | c } 
 \toprule
 \textbf{Method} & \textbf{Fid.(\%)} $\downarrow$ & \textbf{Size }$\downarrow$ \\ [0.5ex] 
 \midrule
 \textsc{Random} & 100 & NULL \\
 \cfg & 100 & NULL \\
  \cff  & 60 & 13.7 +- 16.98 \\
 \name (transductive) & \textbf{53.50} & \textbf{4.72 $\pm$ 4.38} \\
 \bottomrule
\end{tabular}}}\quad
\subfloat[Inductive]{
\label{tab:inductive_all_amazon}
\scalebox{0.85}{
\begin{tabular}{ c | c | c } 
 \toprule
 \textbf{Method} & \textbf{Fidelity(\%)} $\downarrow$ & \textbf{Size }$\downarrow$ \\ [0.5ex] 
 \midrule
  \pgexp & 100 & NULL \\
 \gem & 100 & NULL \\
 \name (inductive) & \textbf{93.00} & \textbf{6.60 $\pm$ 2.87} \\
 \bottomrule
\end{tabular}}}
\caption{ Results for (a) transductive and (b) inductive methods on the Amazon dataset. ``NULL" denotes that the method could not produce a counterfactual.}
\vspace{-0.30in}
\end{table}

\begin{table}[t]
\vspace{0.004in}
\subfloat[Transductive]{
\label{tab:transductive_ogbn}
\scalebox{0.85}{
\begin{tabular}{ c | c | c } 
 \toprule
 \textbf{Method} & \textbf{Fid.(\%)} $\downarrow$ & \textbf{Size }$\downarrow$ \\ [0.5ex] 
 \midrule
 \cfg & DNS & DNS \\
  \cff  & DNS & DNS \\
 \name (transductive) & \textbf{0} & \textbf{1.00 $\pm$ 0.00} \\
 \bottomrule
\end{tabular}}}\quad
\subfloat[Inductive]{
\label{tab:inductive_ogbn}
\scalebox{0.85}{
\begin{tabular}{ c | c | c } 
 \toprule
 \textbf{Method} & \textbf{Fidelity(\%)} $\downarrow$ & \textbf{Size }$\downarrow$ \\ [0.5ex] 
 \midrule
  \pgexp & 95.50 &  4 \\
 \gem & \textsc{DNS} & \textsc{DNS} \\
 \name (inductive) & \textbf{78.7} & \textbf{3.11} \textpm 3.04 \\
 \bottomrule
\end{tabular}}}
\caption{ Results for (a) transductive and (b) inductive methods on the ogbn-arxiv dataset. ``\textsc{DNS}" denotes that the method could not produce a counterfactual as it did not scale. Refer to App.~\ref{app:results_ogbn} that details reasons on why these baselines failed to scale on ogbn-arxiv."}
\vspace{-0.30in}
\end{table}

In Tables \ref{tab:transductive_all_amazon} and \ref{tab:inductive_all_amazon}, we present the results.
 Consistent with the performance on benchmark datasets, \name continues to outperform all the baselines almost in both transductive and inductive settings. We note that most of the baselines failed to produce counterfactuals in Amazon. In ogbn-arxiv, on the other hand, all baselines except \pgexp fails to scale; they crash with out-of-memory exception. In contrast, \name produces promising performance with the transductive version achieving $0\%$ 
 fidelity. 

\vspace{-0.05in}

\vspace{-0.10in}
\subsection{Efficiency}
\label{sec:efficiency}
\vspace{-0.10in}
Table \ref{table:running_time} presents the inference times of various algorithms. First, the inductive methods (\name, \pgexp and \gem) are much faster than the others. Between the inductive methods, \pgexp is the fastest. \name-inductive is slower since the search space for \name is larger due to accounting for both edge additions and deletions. 
 Second, \name-inductive is up to $79$ times faster than the transductive methods such as \cfg and \cff. This speed-up is a result of only doing forward passes through the neural policy network, whereas, transductive methods learn the model parameters on each node separately. Even the transductive version of \name is faster than the other transductive methods for Tree-Cycles and Tree-Grid. 

 \textbf{Scalability against graph size:} Table~\ref{table:scalable_exp} presents the inference time per node across all datasets. We observe that \name scales to million-sized networks such as ogbn-arxiv. We observe that the growth of the running time is closely correlated with the neighbourhood density, i.e., the average degree of the graph, and not the graph size. In a \gnn with $\ell$ layers, only the $\ell$-hop neighborhood of the target node matters.
\begin{table}[t]
\vspace{-0.40in}
\centering
\subfloat[Efficiency]{
\label{table:running_time}
\scalebox{0.65}{
\begin{tabular}{ l|ccc } 
 \toprule
 \textbf{Method} & \textbf{Tree-Cycles} & \textbf{Tree-Grid} & \textbf{BA-Shapes}\\
\midrule
\textbf{\pgexp} & 0.41 & 0.62 & 0.38 \\ 
 \textbf{\gem} & 0.16 & 0.73 & 8.64 \\ 
 \textbf{\cfx} & 1295.66 & 2382.51 & 3964.36\\
 \textbf{\cff} & 165.56 & 249.92 & 2565.87\\ 		
 \textbf{\name (ind.)} & 4.36 & 17.64 & 68.33\\ 
 \textbf{\name (trans.)} & 66.08 & 331.58 & 6546.48\\ 
 \bottomrule
\end{tabular}}}
\subfloat[Scalability]{
\label{table:scalable_exp}
\scriptsize
\centering
\begin{tabular}{ l|cccc } 
 \toprule
 \textbf{Dataset} & \textbf{\#Nodes} & \textbf{\#Edges} & \textbf{Avg. degree} & \textbf{Time/node (ms)}\\
\midrule
\textbf{Tree-Cycles} & 871 & 1,950 & 2.23 & 60.56 \\ 
 \textbf{Tree-Grid}  & 1,231 & 3,410 & 2.77 & 13.67 \\ 
 \textbf{BA-Shapes}  & 700 & 4,100 & 5.86 & 89.91 \\
\textbf{ogbn-arxiv} & 169,343 &1,166,243 & 6.89 &   353.43 \\ 
 \textbf{Amazon-photos} & 7,487 & 119,043 & 15.90 & 5242.32 \\ 		 
 \bottomrule
\end{tabular}}
\caption{(a) Running times (in seconds) of each algorithm on entire test set. (b) Scalability against various graph properties.}
\end{table}
\vspace{-0.10in}
\subsection{Case Study: Counter-factual Visualization}
\vspace{-0.10in}
In this section, we visually showcase how counter-factual explanations reveal vulnerabilities of \gnns and why edge additions are important. 

\begin{figure}[t]
    \centering
    \subfloat[Tree-Cycles]{
\label{fig:tree_cycle_cf_visualization}
    \frame{\includegraphics[width=2.5in]{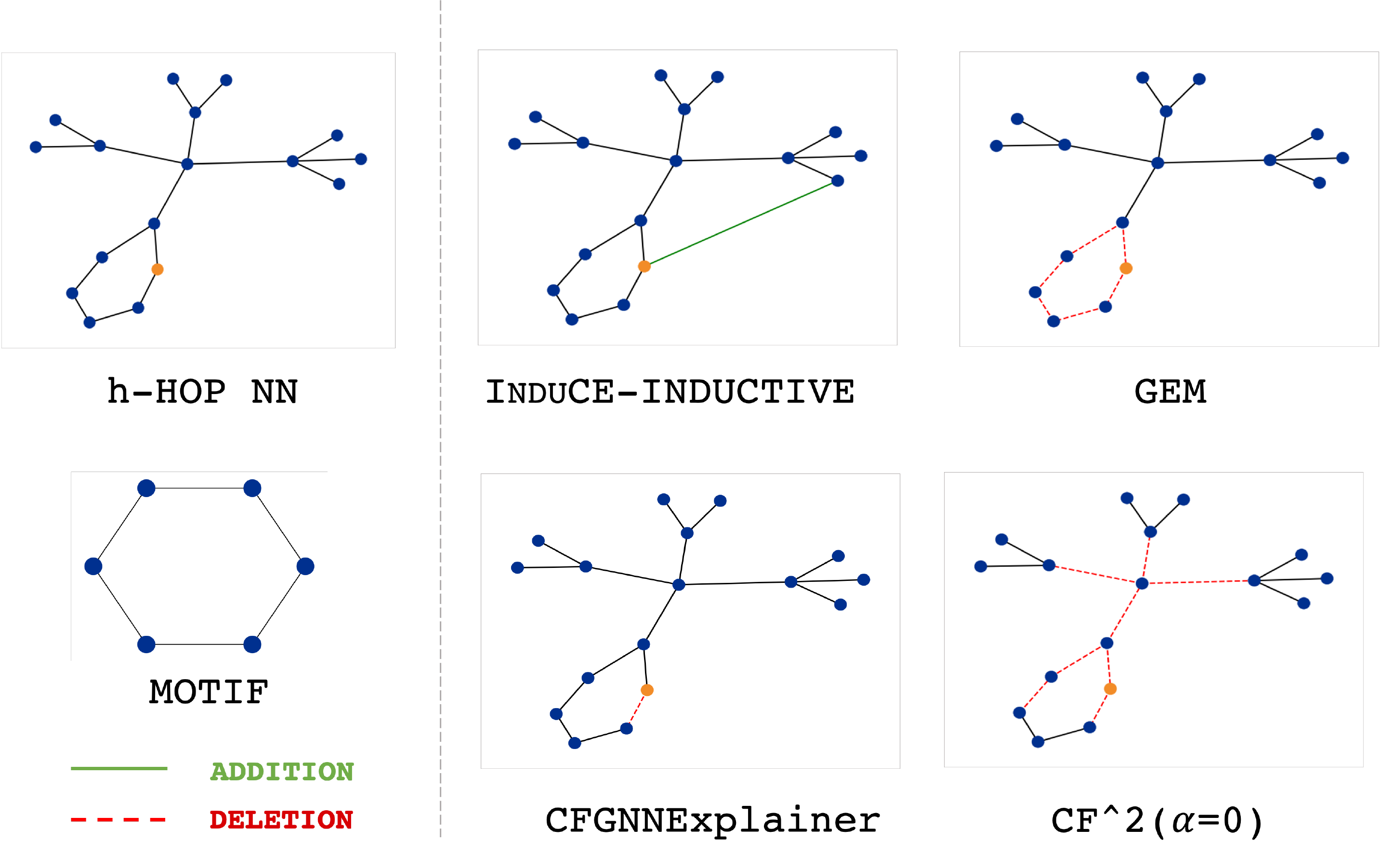}}}
    \hfill\subfloat[Impact of additions]{
    \label{fig:examples_correct_induce}
    \frame{
    \includegraphics[width=2.4in, height=1.55in]{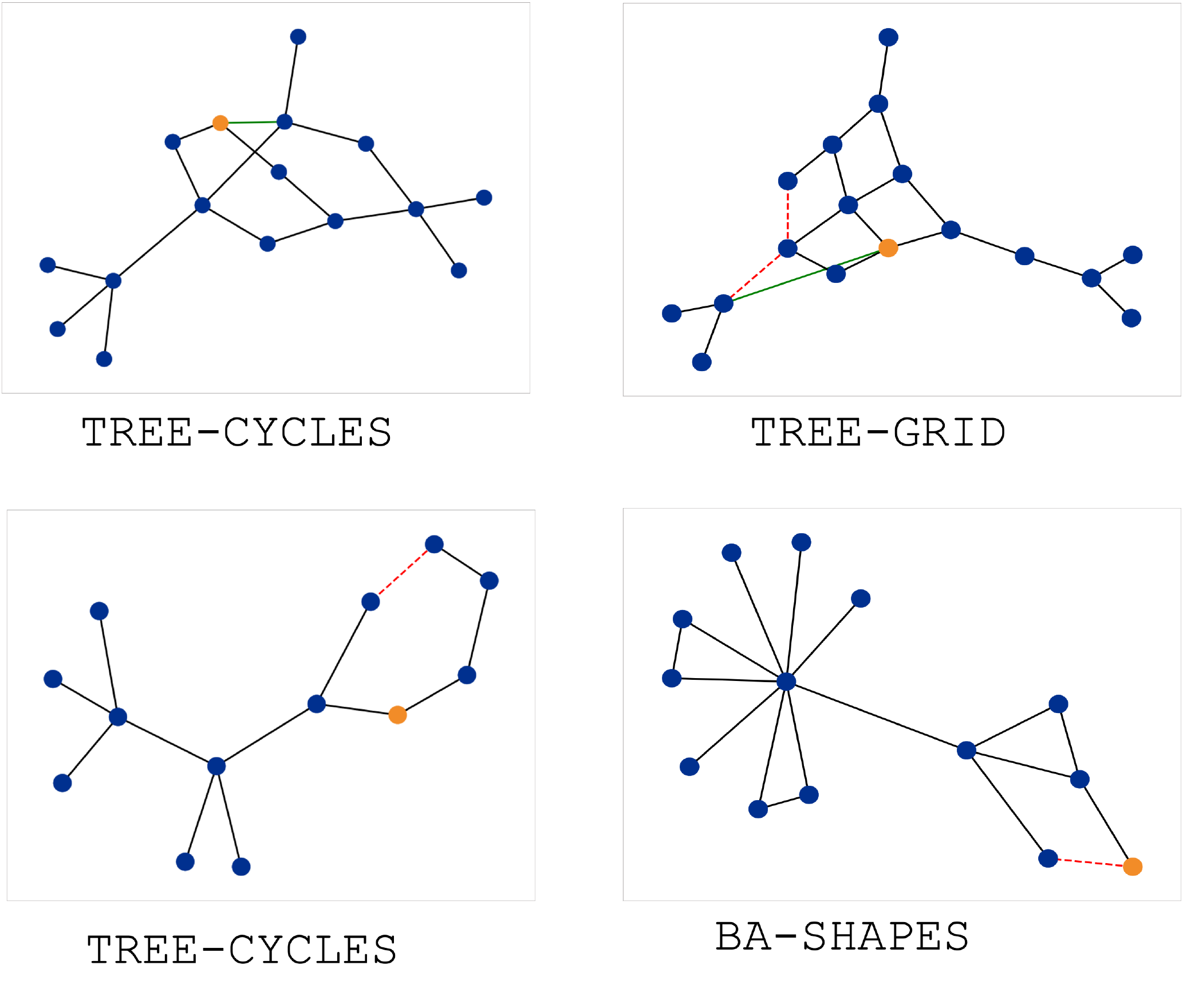}
    }
    }
    \vspace{-0.1in}
    \caption{Visualization of counterfactual explanations for the same node (orange) produced by different methods. Semantically, the node label should flip if it is not a part of the motif. \textbf{(a)} Counterfactual explanation for Tree-Cycles Dataset, \textbf{(b)} Counterfactuals predicted by \name.}
\end{figure}

\textbf{Revealing \gnn vulnerabilities:} 
A sample counterfactual explanation by various algorithms on Tree-Cycles dataset is provided in Fig.~\ref{fig:tree_cycle_cf_visualization}. The target node is part of a motif (6-cycle) and therefore the expected counter-factual explanation is to make it a non-member of a 6-cycle. \cfg correctly finds on such explanation by deleting an edge. Both \gem and \cff recommend a much larger explanation than necessary. In contrast, \name adds an edge. More interestingly, the target node continues to remain part of the motif. This uncovers a limitation of \gnn since it falsely classifies the target node as a non-motif node although it is not. Furthermore, this limitation is uncovered only since \name can add edges. Similar observations in other datasets are available in Figs.~\ref{fig:tree_grid_cf_visualization}-\ref{fig:ba_shapes_cf_visualization} in Appendix.
\looseness=-1

\textbf{Impact of additions:} In Fig.~\ref{fig:examples_correct_induce} top-left, we share an example where \name flips the label of a target node (orange) by making it part of the 6-node cycle motif through edge addition. Since baseline strategies are only capable of deletes, they fail to flip the label of such nodes. Further in the top-right (Tree-Grid) example, we see \name breaks the grid motif and connects the target node (orange) to a non-motif neighbour, hence colluding its embeddings and flipping its label. 
These examples showcase the importance of edge addition 
 to intuitively explain how the black-box \gnn works. 

\vspace{-0.10in}

\vspace{-0.05in}
\section{Conclusion}
\vspace{-0.10in}
The ability to explain predictions is critical towards making a model trustworthy. In this work, we proposed \name to understand \gnns via counterfactual reasoning for the node classification task. While several algorithms in the literature produce counterfactual explanations of \gnns, they suffer from restricted counterfactual space exploration and transductivity. \name provides a boost to counterfactual analysis on \gnns by unleashing the power of edge additions and inductively predicting explanations on unseen nodes. The proposed features not only lead to better explanations but also provide a significant speed-up allowing \name to perform counterfactual analysis at scale.

\textbf{Limitations:} \name performs counter-factual reasoning by perturbing only the topological space. In future, we will consider characterization of the node feature space and explore the joint combinatorial space of topology and features.  

\bibliographystyle{plain}
\bibliography{arxiv_main}

\clearpage
\section*{Appendix}
\label{sec:paper_appendix}

\appendix
\renewcommand{\thesubsection}{\Alph{subsection}}
\renewcommand{\thefigure}{\Alph{figure}}
\renewcommand{\thetable}{\Alph{table}}

\begin{table*}[t]
\footnotesize
\scalebox{0.90}{
\begin{tabular}{ l||c|c|c|c|c} 
\toprule
 \textbf{Method} & \textbf{Explainability Paradigm} & \textbf{Additions}  & \textbf{Deletions} & \textbf{Inductive} & \textbf{\# explainers required} \\ 
\midrule
\textbf{\textsc{GNNExplainer}}\cite{gnnexplainer} & Factual & \xmark & \cmark & \xmark & $N$\\
\textbf{\textsc{GraphMask}}\cite{schlichtkrull2020interpreting} & Factual & \xmark & \cmark & \xmark & $N$\\
\textbf{\textsc{Causal Screening}}\cite{wang2021causal} & Factual & \xmark & \cmark & \xmark & $N$\\
\textbf{\textsc{SubgraphX}}\cite{subgraphx} & Factual & \xmark & \cmark & \xmark & $N$\\ 
\textbf{\textsc{PGM-Explainer}}\cite{vu2020pgm} & Factual & \xmark & \cmark & \xmark & $N$\\ 
\textbf{\textsc{PGExplainer}}\cite{pgexplainer} & Factual & \xmark & \cmark & \cmark & $1$\\
\textbf{\textsc{GEM}}\cite{gem} & Counterfactual & \xmark & \cmark & \cmark & $1$\\
\textbf{\textsc{CF-GNNExplainer}}\cite{Cfgnnexplainer} & Counterfactual & \xmark & \cmark & \xmark  & $N$\\
\textbf{\textsc{CF$^2$}}\cite{cff} & Counterfactual + Factual & \xmark & \cmark & \xmark & $N$ \\
\textbf{\textsc{InduCE (Ours)}} & Counterfactual & \cmark & \cmark & \cmark & $1$\\
\bottomrule
\end{tabular}}
\caption{\textbf{Comparison on properties of common perturbation-based GNN explainers. The last column shows the number of required explainers for a graph with $N$ nodes.}}
\label{table:baseline_comparison}
\vspace{-0.05in}
\end{table*}

\begin{table}[t]
\centering
\small
\begin{tabular}{ c|cccc } 
\toprule
\textbf{Dataset} & \textbf{k-hop}& \textbf{\#additions}& \textbf{\#deletions}& \textbf{ratio (\#additions/\#deletions)}\\
\midrule
 \textbf{Tree-cycles} & 4 & 18313 & 1105 & 16.57 \\
 \textbf{Tree-grid} & 4 & 98942 & 3960 & 24.98 \\
 \textbf{BA-shapes} & 4 & 3140886 & 47580 & 66.01 \\
 \textbf{Amazon-Photos} & 4 & 12144800 & 432160 &	28.10 \\
 \textbf{ogbg-arxiv} & 3 & 354496362	& 524330 &	676.09 \\
\bottomrule
\end{tabular}
\caption{Ratio of no. of additions to deletions of datasets.}
\label{tab:actionspace}
\end{table}

\begin{figure}[t]
\vspace{-0.4in}
    \centering
    \subfloat[Tree-Grid]{
      \label{fig:tree_grid_cf_visualization}
    \frame{\includegraphics[width=2.7in]{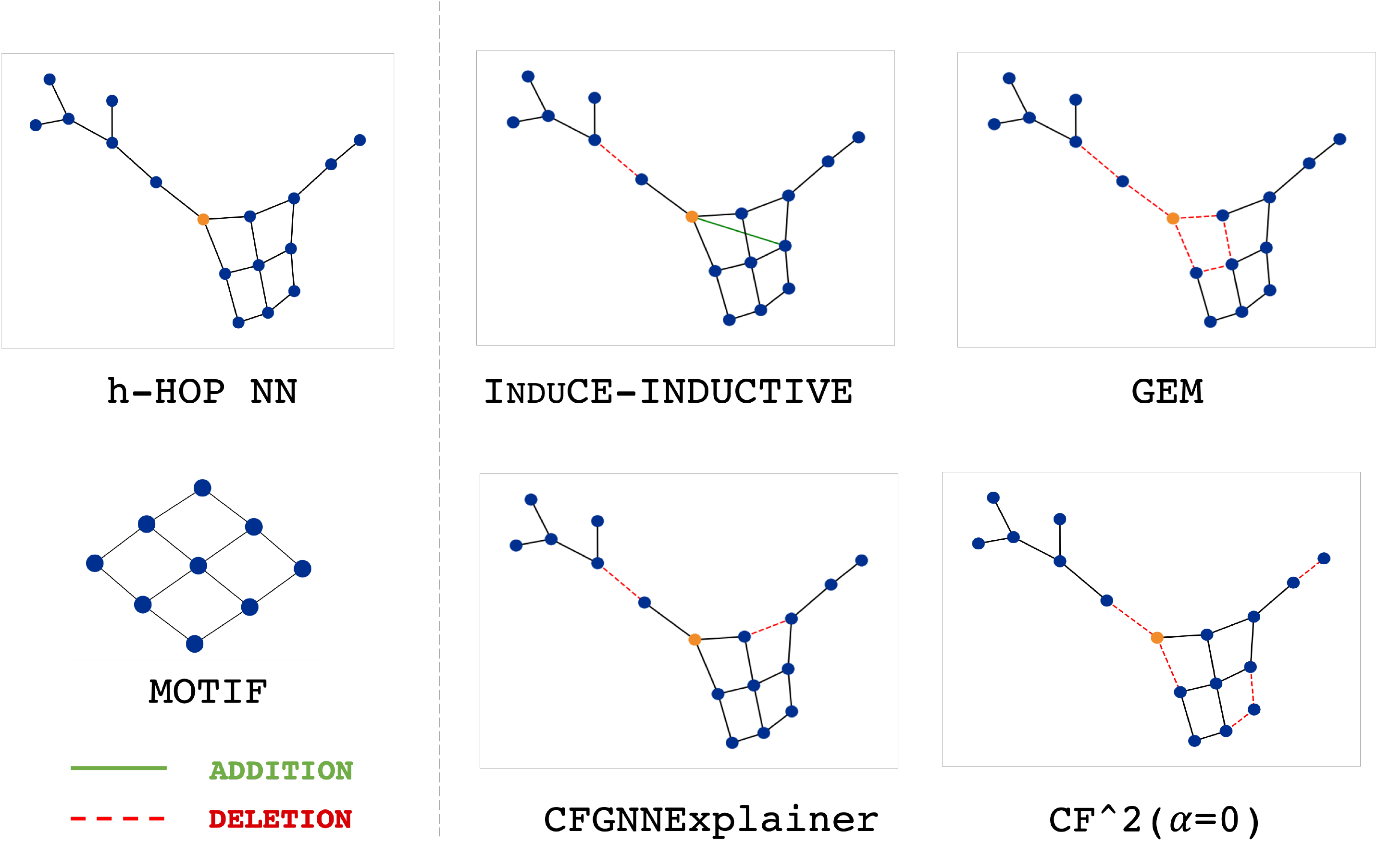}}}
    \subfloat[BA-Shapes]{
    \label{fig:ba_shapes_cf_visualization}
    \frame{\includegraphics[width=2.7in]{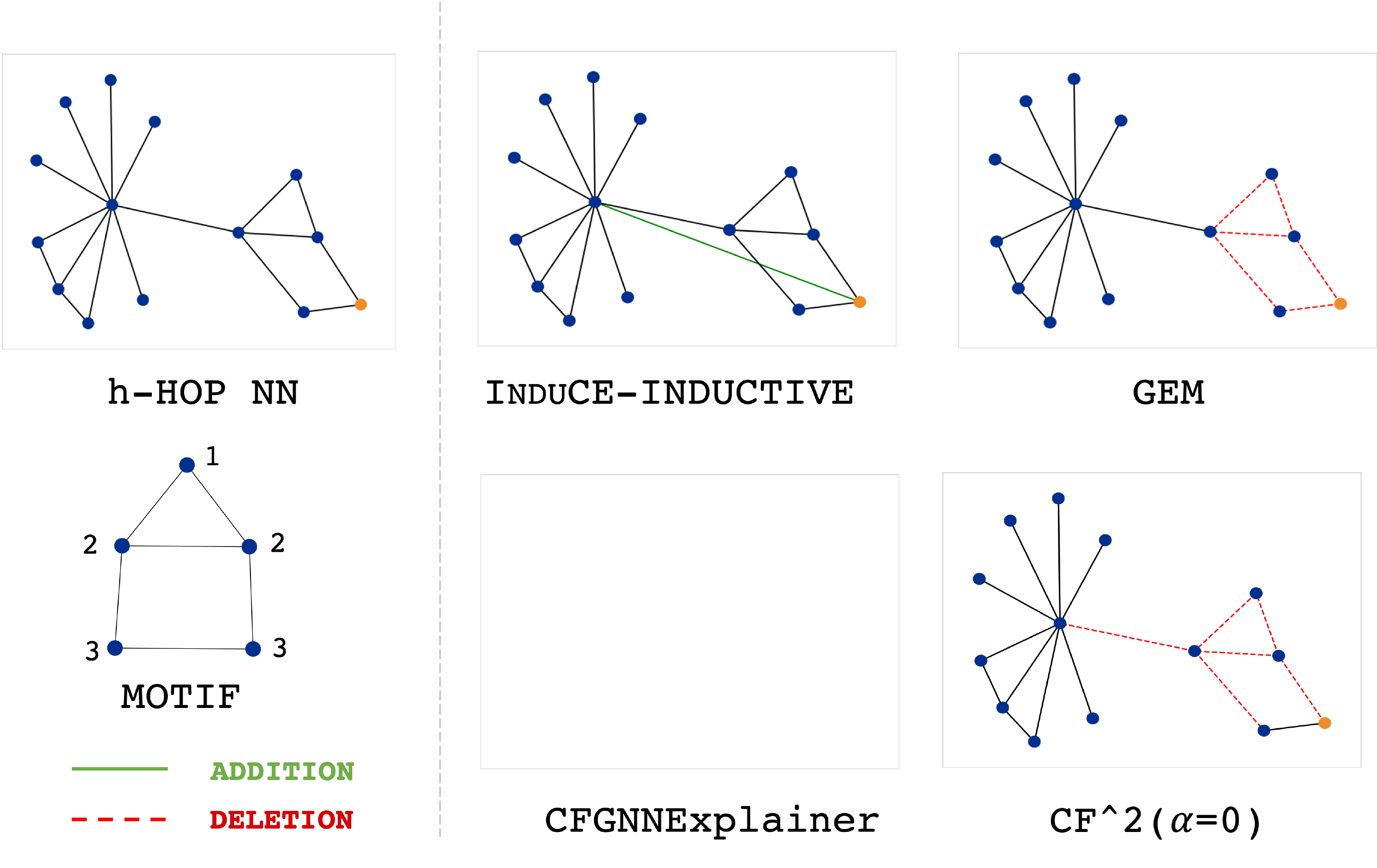}}
    }
    
    \vspace{-0.1in}
    \caption{Visualization of counterfactual explanations for the same node (orange) produced by different methods. Semantically, the node label should flip if it is not a part of the motif, i.e., \textbf{(a)} $3\times3$-grid, and the \textbf{(b)} the house respectively. \cfg is unable to find a counterfactual in \textbf{(b)}.}
    \vspace{-0.05in}
\end{figure}

\renewcommand{\thesubsection}{\Alph{subsection}}
\begin{algorithm}[t]
\caption{Training pipeline of \name.}
\label{alg:induce}
{\scriptsize
\begin{flushleft}
\textbf{Input:} Graph $\CG$, \gnn $\Phi$, Train set $\mathcal{V}_{tr}$, perturbation budget $\delta$, number of episodes $M$\\
\textbf{Output:} Policy $\Pi$\\
\end{flushleft}
\begin{algorithmic}[1]
\State $\mathcal{V}_{batch}\ \leftarrow\ \{\mathcal{V}_1, \mathcal{V}_2\,\dots,\mathcal{V}_B|\cup_{i=1}^{\mathbb{B}}\mathcal{V}_i=\mathcal{V}_{tr}\}$ \Comment{Random partitioning instances of $\mathcal{V}_{tr}$ into $B$ batches}
\State $\Pi\leftarrow$ initialize with random parameters
\ForAll{$e\ \in[1,M]$}
\ForAll{$\mathcal{V}_{b}\in\mathcal{V}_{batch}$}
\ForAll{$v\in\mathcal{V}_{b}$}
\State $t\leftarrow 0$
\While {$L_\Phi(\mathcal{G}^0,v)=L_\Phi(\mathcal{G}^t,v) \And t < \delta$}
\State compute $\mathcal{S}_v^t$
\State $a^t\leftarrow\ \textrm{sample from}\  \Pi(\mathcal{S}_v^t, \mathcal{P}^t)$
\State $\mathcal{G}_v^{t+1} \leftarrow\ \textrm{perturb}\ \mathcal{G}_v^t\ \textrm{with edge}\ a^t$
\State $\mathcal{R}_v^{t} \leftarrow\ \textrm{compute reward using Eq.10}$
\State $\mathcal{R}_{dis,v}^{t} \leftarrow\ \textrm{compute discounted rewards using Eq.\ref{eq:dis_reward}}$
\State $\mathcal{\tilde{R}}_{dis,v}^{t} \leftarrow\ \textrm{normalize discounted rewards using Eq.\ref{eq:norm_reward}}$
\State $t\leftarrow t+1$
\EndWhile
\EndFor
\State Backpropagate to minimize loss using Eq.~\ref{eq:loss_batched}
\EndFor
\EndFor
\State \textbf{Return} $\Pi$
\end{algorithmic}}
\end{algorithm}

\begin{algorithm}[t]
\caption{Test pipeline of \name.}
\label{alg:induce-test}
{\footnotesize
\begin{flushleft}
\textbf{Input:} Graph $\CG$, \gnn $\Phi$, Test set $\mathcal{V}_{test}$, maximum perturbation budget $\delta$\\
\textbf{Output:} Counterfactual explanations $\textsc{CF}$\\
\end{flushleft}
\begin{algorithmic}[1]
\State $CF=\phi$
\State $\Pi\leftarrow$parameters of pre-trained policy  
\ForAll{$v\in\mathcal{V}_{test}$}
\State $Exp\leftarrow\emptyset$
\State $t\leftarrow 0$
\While {$L_\Phi(\mathcal{G}^0,v)=L_\Phi(\mathcal{G}^t,v) \And t < \delta$}
\State compute $\mathcal{S}_v^t$
\State $a^*\leftarrow\arg\max_{a\in\mathcal{P}^t}\Pi(\mathcal{S}_v^t, \mathcal{P}^t)$
\State $\mathcal{G}_v^{t+1} \leftarrow\ \textrm{perturb}\ \mathcal{G}_v^t\ \textrm{with edge}\ a^*$
\State $t\leftarrow t+1$
\EndWhile
\If{$L_\Phi(\mathcal{G}^0,v)\neq L_\Phi(\mathcal{G}^t,v)$}
\State $CF=CF\cup Exp$
\EndIf
\EndFor
\State \textbf{Return} $\textsc{CF}$
\end{algorithmic}}
\end{algorithm}

\subsection{Proof of Theorem \ref{thm:nphard}}
\label{sec:app_proof_nphard}

\textsc{Proof.} To prove NP-hardness of the problem we reduce it from the classical \textit{set cover} problem.
\begin{defn}[Set Cover \cite{feige1998threshold}]
Given a collection of subsets $\mathcal{S}=\{S_{1},\cdots,S_{m}\}$ from a universe of items $U=\{u_{1},\cdots,u_{n}\}$  identify the smallest collection of subsets $\mathbb{\mathcal{A}^*}\subseteq \mathcal{S}$ covering the set $U$, i.e., 
\begin{equation}
\mathcal{A}^*=\arg \min_{\mid \mathcal{A} \mid, \mathcal{A}\subseteq \mathcal{S}}\bigcup_{\forall S_i\in \mathbb{\mathcal{A}}} S_i= U
\end{equation}
\end{defn}

We show that given any instance of a set cover problem $\langle \mathcal{S}, U\rangle$, it can be mapped to Prob.~\ref{prb:cf}. Specifically, we construct a graph $\CG=(\CV,\CE)$, where $\CV=N\cup\mathcal{S}\cup U$. Here, $N$ is an arbitrary set of nodes. In addition, we have a node corresponding to each set $S\in\mathcal{S}$ and each item $u\in U$. There is an edge between two nodes $v_i,v_j\in\CV$ if $v_i$ corresponds to some set $S\in\mathcal{S}$, $v_j$ corresponds to item $u\in U$, and $u\in S$. There are no edges among nodes in $N$. The \gnn $\Phi$ predicts the label of any node $v\in N$ as $1$ if all nodes from $U$ are reachable from $v$, otherwise $0$. Furthermore, let the set of allowed perturbations be $\CV_c=\{(v_i,v_j)\mid v_i\in N, v_j\in\mathcal{S}\}$. Given any $v\in N$, with $L_{\Phi}(\CG,v)=0$, the counterfactual reasoner therefore needs to identify the minimum number of edges to add so that all nodes from $U$ are reachable from $v$ through some nodes in $\mathcal{S}$. 
 
 With this construction, only edge additions are allowed. Now it is easy to see that the smallest edge set flipping the label of $v$ corresponds to connecting $v$ to nodes in $\mathcal{A}^*$, where $\mathcal{A}^*$ is the solution for the set cover problem. $\hfill \square$.

\subsection{Discounted Rewards}
\label{app:loss}
The objective of the policy $\Pi$ is to find minimal counterfactual explanations for \gnns using the reward function mentioned in Eq 10. However, we can observe that in that equation, the marginal reward at each step is given equal weight. In our case, we want the immediate rewards to have higher weight over the rewards encountered later on in the perturbation trajectory $\mathcal{T}_v$ in order to penalize larger counterfactual size, thus we use discounted rewards (Eq \ref{eq:dis_reward}) with $\gamma$ being the discount factor to achieve the objective. Since we want minimal explanation size, we use small values of $\gamma$ (Refer App.  \ref{app:parameters}).

\begin{equation}
\label{eq:dis_reward}
    \mathcal{R}^{t}_{dis,v}(a_t)=\sum_{i=0}^{\delta} \gamma^{i}\mathcal{R}_v^{t+i+1}(a_t)
\end{equation}
here $\delta$ is the maximum perturbation budget.\\
One limitation of policy gradient is high variance caused by the scale of rewards. A common way to reduce variance is to subtract a baseline, $b(S_v^t)$ such that it does not induce bias in the policy gradient. A simple baseline can be the mean of the discounted rewards, so that we train the policy to pick trajectories that give rewards better than the average rewards. We also normalize the discounted reward further by dividing with the standard deviation. 
\begin{equation}
\label{eq:norm_reward}
    \mathcal{\tilde{R}}^{t}_{dis,v}(a_t) = \frac{\mathcal{R}^{t}_{dis,v}(a_t) - \mathcal{\bar{R}}^{t}_{dis,v}(a_t) }{\max( \sigma{(\mathcal{R}^{t}_{dis,v}(a_t)),\ c)} }
\end{equation}
where $\sigma{(\mathcal{R}^{t}_{dis})}$ is the standard deviation, $c$ is a constant. The optimized version of the loss function(Eq \ref{eq:loss}) is in App. \ref{app:batching}.

\subsection{Batching}
\label{app:batching}
Equation \ref{eq:loss} takes an average gradient over all examples in the training set. This setting may lead to over-smoothing of the gradients and hence induce difficulty in training. To counter this issue, we performed batching of node instances, and back-propagated with the average gradients computed on nodes in the batch. Thus we optimize our policy on batches of nodes, and use normalized discounted rewards in the loss function(Refer Eq. \ref{eq:loss_batched}).

\begin{equation}
\scriptsize
\label{eq:loss_batched}
\mathcal{J}(\Pi) =-\frac{1}{\CV_{batch}}\left(\sum_{\forall v\in\CV_{batch}}\left(\sum_{t=0}^{\lvert \CT_v \rvert}\log{p^t_{a,v}} \mathcal{\tilde{R}}^t_{dis,v}(a_t)+\eta Ent(\CP^t_v)\right)\right)
\end{equation}
\subsection{\name : Algorithm}
\label{app:induce_algorithm}
The training and test pipelines of \name are provided in Alg. \ref{alg:induce} and Alg. \ref{alg:induce-test} respectively. As evident from Alg. \ref{alg:induce-test}, using \name we can train once and test on unseen nodes using just a forward pass through the policy network which makes it more efficient than the transductive baselines (recall Table \ref{table:running_time}).

\begin{table}[t]
\centering
\small
\begin{tabular}{ c|cc } 
\toprule
\textbf{Dataset} & \textbf{Train accuracy(\%) }& \textbf{Test Accuracy(\%)}\\
\midrule
 \textbf{Tree-cycles} & 91.23 & 90.86\\
 \textbf{Tree-grid} & 84.34 & 87.44\\
 \textbf{BA-shapes} & 96.61 & 98.57\\
 \textbf{Amazon-Photos} & 89.59 & 88.75\\
 \textbf{ogbn-arxiv} & 71.59 & 55.07\\
\bottomrule
\end{tabular}
\caption{Accuracy of \gnn $\Phi$ on the three benchmark datasets.}
\label{table:model_accuracy}
\end{table}

\subsection{Complexity of \name}
\label{sec:app_complexity}


\textbf{Train-Phase:}
Training the policy network involves four key steps, i.e., compute the state, forward pass through the policy network, sample and take action, and compute the marginal reward of the action. Among the above, state computation and performing the action take $\mathcal{O}(1)$ time. Time taken by a forward pass through the policy network involves a combination of computing node embeddings using \gat and computing a score for each action in the action space $\mathcal{P}^t$ through the \MLP. Forward pass through the \gat takes $\mathcal{O}(K(\mathcal{|V|}h_ih_d + \mathcal{|E|}h_d))$ ~\citep{gat} where $K$, $h_i$ and $h_d$ are the number of \gat layers, input and hidden dimensions respectively. Action score computation using \MLP takes $\mathcal{O}(|\mathcal{P}^t|h_m(h_d + (J-2)h_m + 1)))$, $J$ and $h_m$ are number of \MLP layers and hidden dimensions. Computing the reward function involves taking a forward pass through a \textsc{GCN} which takes $\mathcal{O}(L|\mathcal{E}|h_ih_d\mathcal{|C|})$ time, here $L$ and $\mathcal{|C|}$ are the number of layers and number of classes respectively. The above steps are repeated for $M$ episodes for each node in training set $V_{tr}$ until a maximum of $\delta$ time steps. Combining the above costs, treating $J$, $K$, $L$, $h_i$, $h_d$, $h_m$ and $\mathcal{|C|}$ as fixed constants which have small values, and with the knowledge that $|\mathcal{P}^t|=\mathcal{O}(|\mathcal{V}|+|\mathcal{E}|)$ (refer Eq. \ref{eq:perturbations}), the complexity with respect to the input parameters reduces to $\mathcal{O}(|\mathcal{V}_{tr}|M\delta(\mathcal{|V|} + \mathcal{|E|}))$. \\
\\
\textbf{Test-Phase: }The test phase involves state computation, forward pass through the policy network, and performing the action with highest probability for a maximum of  $\delta$ time-steps for every node in the test set $\mathcal{V}_{test}$. Therefore following the discussion in the training phase, the time complexity of test phase is $\mathcal{O}(|\mathcal{V}|_{test}|\delta(\mathcal{|V|} + \mathcal{|E|}))$.

Detailed algorithm of \name is given in App. \ref{app:induce_algorithm}.

\subsection{Experimental Setup}
\label{app:setup}
All reported experiments are conducted on an NVIDIA DGX Station with four V100 GPU cards having 128GB GPU memory, 256GB RAM, and a 20 core Intel Xeon E5-2698 v4 2.2 Ghz CPU running in Ubuntu 18.04.

\subsubsection{Benchmark datasets}
\label{app:datasets}
\begin{itemize}
    \item \textbf{BA-SHAPES: }The base graph is a Barabasi-Albert (BA) graph. The motifs are \textbf{house-shaped} structures made up of 5 nodes (Refer Figure \ref{fig:ba_shapes_cf_visualization}). Non-motif nodes are assigned class $0$, while nodes at the top, middle, and bottom of the motif are assigned classes $1$, $2$, and $3$, respectively. 
    \item \textbf{TREE-CYCLES: }The base graph is a binary tree with \textbf{6-node cycles} used as motifs (Refer Figure \ref{fig:tree_cycle_cf_visualization}). The motifs are connected to random nodes in the tree. Non-motif nodes are labelled 0, while the motif nodes are labelled 1. 
    \item \textbf{TREE-GRID: }The base graph is a binary tree and the motif is a \textbf{$\mathbf{3\times3}$ grid} connected to random tree nodes (Refer Figure \ref{fig:tree_grid_cf_visualization}). Just like tree-cycles dataset binary class labelling has been done. 
\end{itemize}
\subsubsection{Baselines}
\label{app:baselines}
\begin{itemize}
    \item \textbf{\cfg} \cite{Cfgnnexplainer}: Being a transductive method for counterfactual explanations, it learns a new set of parameters for every node and cannot be used to explain unseen nodes. 
    
    \item \textbf{\cff} \cite{cff}: While being transductive in nature, it combines both counterfactual and factual properties to give an explanation. \cff  tunes the parameter $\alpha$ to weigh the contribution of factual explanations. We compare \cff with $\alpha=0$ where it becomes as a counterfactual explainer. 
    \item \textbf{\gem} \cite{gem}: This is inductive by nature, however, it only considers edge deletions. It has a limitation that it learns a counterfactual explanation model where the number of perturbations is fixed, i.e., it does not minimize the number of perturbations with the sole focus on flipping the label. We use the default size of $6$ as the perturbation size as recommended by the authors. 

Note that \gem is not extendable to include edge additions. Specifically, GEM has a distillation process that generates the ground truth. Distillation involves removing every edge in a node's neighbourhood iteratively and seeing its effect on the loss. The deletions are then sorted based on their effect on the loss. The top-$k$ edges ($k$ is user-specified) are used as the distilled ground truth. The explainer is later trained to generate graphs that are the same as the distilled ground truth. To extend this process for additions, the number of possible edge edits is significantly higher and the iterative process of GEM to create the distilled ground truth does not scale. In addition, it is also unclear how to set $k$ in the presence of additions.
    
    \item \textbf{\pgexp} \cite{pgexplainer}: This method is also inductive and only considers edge deletions. It is a factual explainability method and requires a fixed explanation size as a hyper-parameter. We use the default size of $6$ as the perturbation size as recommended by the authors for the benchmark datasets. We also use size $6$ and $4$ for Amazon-Photos and ogbn-arxiv, respectively.
    \item \textbf{\textsc{Random: }} We use the same baseline as used in \cite{Cfgnnexplainer}. It makes the choices of deleting an edge randomly by generating a random subgraph mask for the $h$-hop neighbourhood of the node and perturbing it.  
    \end{itemize}

\subsubsection{Training and Parameters}
\label{app:parameters}

\paragraph{Counter-factual task:} We provide a node that is part of a motif to the counterfactual explainer, and the the task is to flip its label by recommending changes in the graph. All nodes that are part of a motif, are given a specific label and non-motif nodes are given a different label. Since the nodes are always chosen from motifs, the explanation is the motif itself.  This setup is identical to \cff and \cfg.


\paragraph{The \gnn model $\Phi$: } 
We use the same \gnn model used in \cfg and \cff. Specifically, it is a Graph Convolutional Networks trained on each of the datasets. Each model has 3 graph convolutional layers, with 20, 128 and 256 hidden dimensions for the benchmarking datasets, Amazon-photos and ogbn-arxiv respectively. The non-linearity used is \textit{relu} for the first two layers and \textit{log \softmax} after the last layer of GCN. The learning rate is 0.01. The train and test data are divided in the ratio 80:20 for benchmark datasets. \newapp{For ogbn-arxiv, we use the standard splits provided in the ogb package. In our experiments, we use a scaled-down version of the Amazon-Photos dataset. We choose one random node as the central node and took its $3-$hop neighbourhood in our dataset. Amazon Photos has an average degree of 13, hence, the $3-$hop neighborhood covers a reasonable distribution of class labels. We split the nodes of this subgraph in the ratio of $80:20$ for train and test sets. 
The accuracy of the \gnn model  $\Phi$ for each dataset is mentioned in Table \ref{table:model_accuracy}}.

\noindent\textbf{Training, Inference and Parameters: } For \name and \gem, we use a \sv{train/evaluation split of 80/20} \newapp{on the benchmark and \newapp{the Amazon-Photos} datasets. For ogbn-arxiv, we train on 10 random examples per class, and sample 1000 random nodes as the test dataset. We make sure that the test and train sets are disjoint}. The evaluation set for all techniques are identical. For \gem and \name, the train set is identical. Since \cfg and \cff are transductive, only the evaluation set is used for them where they learn a node-specific parameter set. The same happens on the transductive version of \name. 

\paragraph{Parameters settings:}
We use $h=4$ because extracting the 4-hop neighbourhood as the subgraph ensured that we preserve the black-box model's accuracy. We use $\beta=0.5$ so as to give equal weight to the predict loss and distance loss (see Eq. 10). We use different values of $\gamma \in \{0.4, 0.6\}$ and find the best performance at $\gamma=0.4$ for the inductive setting and $\gamma=0.6$ for the transductive setting with a maximum perturbation budget $\delta = 15$.  We use maximum number of episodes $\mathbb{M}=80, 500, 500$ for BA-shapes, Tree-cycles and Tree-grid respectively. We use GAT as the GNN of choice for the policy network. For the policy network, we use 3 GAT layers, 2 fully connected MLP layers, 16 hidden dimension, a learning rate of $.0003$ and LeakyReLU with negative slope $0.1$ as the activation function. We use three different values for $\eta \in \{0.1, 0.01, 0.001\}$ and $\eta=0.1$ improves the performance due to higher weight for exploration. 

\begin{table*}[h!]
\centering
\scalebox{0.9}{
\scriptsize
\begin{tabular}{ l||c||c||c||c||c } 
\toprule
\textbf{Method} & {\textbf{Sparsity  (Tree-Cycles)}}  & \textbf{Sparsity (Tree-Grid)} & \textbf{Sparsity (BA-Shapes)} & \textbf{Amazon-photos} & \textbf{ogbn-arxiv}\\ 
\midrule
\cfg & \textbf{0.93} & 0.95 & 0.99 & NULL & DNS\\ 	
\cff & 0.52 & 0.59 & 0.92  & 0.99 & DNS\\ 
\name - transductive & 0.92 & \textbf{0.96} & \textbf{0.98} & \textbf{0.99} & \textbf{0.78}\\ 
 \bottomrule
\end{tabular}}
\caption{\textbf{ Comparison of "sparsity" of counterfactuals predicted by transductive methods. \newapp{``NULL'' means the baseline could not find a counterfactual. ``DNS'' means that the baseline did not scale.}}}
\label{table:sparsity_tran}
\vspace{-0.10in}
\end{table*}

\begin{table*}[h!]
\centering
\scalebox{0.9}{
\scriptsize
\begin{tabular}{ l||c||c||c||c||c } 
\toprule
\textbf{Method} & {\textbf{Sparsity (Tree-Cycles)}}  & \textbf{Sparsity (Tree-Grid)} & \textbf{Sparsity (BA-Shapes)} & \textbf{Amazon-photos} & \textbf{ogbn-arxiv}\\ 
\midrule
\pgexp & 0.34 & 0.64 & 0.61 & NULL & \textbf{0.66}\\ 
\gem & 0.54 & 0.77 & 0.88 & NULL & DNS\\ 
\name - inductive & \textbf{0.81} & \textbf{0.83} & \textbf{0.98}  & \textbf{0.99} & 0.64\\ 
\bottomrule
\end{tabular}}
\caption{\textbf{Comparison of "sparsity" of counterfactuals predicted by inductive methods. \newapp{``NULL'' means the baseline could not find a counterfactual. ``DNS'' means that the baseline did not scale.}}}
\label{table:sparsity_ind}
\vspace{-0.10in}
\end{table*}

\begin{table}[b]
\centering
{\footnotesize
\begin{tabular}{ c||ccc } 
 \toprule
 \multirow{2}{*}{\textbf{Policy Variant}} & \multicolumn{3}{c}{\textbf{Tree-Cycles}}\\\cline{2-4} 
 & \textbf{Fid.(\%)} $\downarrow$ & \textbf{Size} $\downarrow$ & \textbf{Acc.(\%)} $\uparrow$\\
 \midrule
\multicolumn{1}{c||}{\centering \textbf{\name-inductive-\textsc{Gcn}}} & \textbf{0} &  2.62 \textpm 1.52 & 78.84 \\
 \multicolumn{1}{c||}{\centering \textbf{\name-inductive-\gat}} & \textbf{0} &  \textbf{1.99 \textpm 1.00} & \textbf{97.47}\\ 	
  \hline 
 \multicolumn{1}{c||}{\centering \textbf{\name-transductive-\textsc{Gcn}}} &  \textbf{0} & \textbf{1.08 \textpm 0.27} & \textbf{96.84}\\
 \multicolumn{1}{c||}{\centering \textbf{\name-transductive-\gat}} & \textbf{0} & \textbf{1.08 \textpm 0.27} & 96.20\\
 \hline
\end{tabular}}
\vspace{0.2cm}
\caption{Importance of attention in the \gnn component of \name.} 
\label{table:policy_variation}
\vspace{-2mm}
\end{table}

\subsection{Additional Results on Sparsity of Counterfactuals}
\label{app:sparsity}
Recall, sparsity is defined as the proportion of edges in $\CN^{l}_v$, i.e., the $\ell$-hop neighbourhood of the target node $v$.  Since counterfactuals are supposed to be minimal, a value close to 1 is desired. We compare \name with its baselines on sparsity in Tables \ref{table:sparsity_tran} and \ref{table:sparsity_ind}. \newapp{We observe that \name produces better or comparable explanations in terms of sparsity. In Table ~\ref{table:sparsity_ind} we observe that the sparsity of \name-inductive is slightly less than \pgexp for the ogbn-arxiv dataset. However, the fidelity of \pgexp is greater than \name-inductive (Recall Table ~\ref{tab:inductive_ogbn}). Thus, when looking at the combined results, one may conclude that \pgexp finds counterfactual explanations for the easier examples and, as a result, has sparser explanations. Similarly, we can interpret the sparsity of \cfg being better than \name-transductive for Tree-Cycles in Table \ref{table:sparsity_tran} as its fidelity is much higher than the latter (Recall Table ~\ref{table:node_class_results_transductive}).]} 



\begin{table*}[t]
\scalebox{0.82}{
\scriptsize
\begin{tabular}{ 
 c||ccc||ccc||ccc } 
 \toprule
 \multirow{2}{*}{\textbf{Method}} & \multicolumn{3}{c||}{\textbf{Tree-Cycles}}  & \multicolumn{3}{c||}{\textbf{Tree-Grid}} & \multicolumn{3}{c}{\textbf{BA-Shapes}}\\\cline{2-10} 
 & \textbf{Fid.(\%)} $\downarrow$ & \textbf{Size} $\downarrow$ & \textbf{Acc.(\%)} $\uparrow$ & \textbf{Fid.(\%)} $\downarrow$ &  \textbf{Size} $\downarrow$ & \textbf{Acc.(\%)} $\uparrow$ & \textbf{Fid.(\%)} $\downarrow$ & \textbf{Size} $\downarrow$ & \textbf{Acc.(\%)} $\uparrow$\\
 \midrule
\multicolumn{1}{p{3.5cm}||}{\centering \textbf{Features only}} & \textbf{0} &  3.12 \textpm 1.96 & 80.05 & \textbf{0} & 4.50 \textpm 3.16 & 73.12 & 18.4 & 3.62 \textpm 3.46 & 70.29\\ 
 \multicolumn{1}{p{3.5cm}||}{\centering \textbf{Features + D}} &  \textbf{0} & 2.56 \textpm 1.73 & 65.74 & \textbf{0} & 3.71 \textpm 2.51 & 84.26 & 9.2 & 4.02 \textpm 4.27 & \textbf{98.89}\\ 
 \multicolumn{1}{p{3.5cm}||}{\centering \textbf{Features + E}}& \textbf{0} & \textbf{2.24 \textpm 1.15} & 72.69 & \textbf{0} & 3.37 \textpm 2.22 & \textbf{94.63} & 68.4 & \textbf{1.25 \textpm 0.83} & 86.25\\ 
\multicolumn{1}{p{3.5cm}||}{\centering \textbf{Features + OH}}& \textbf{0} & 3.19 \textpm 1.83 & 79.70 & 2.3 & 4.10 \textpm 3.13 & 87.17 & 28.9 & 4.63 \textpm 3.41 & 32.75\\ 
\multicolumn{1}{p{3.5cm}||}{\centering \textbf{Features + D + E}}& \textbf{0} & 2.81 \textpm 1.55 & 85.19 & \textbf{0} & \textbf{3.12 \textpm 1.96} &  84.77 & 48.7 & 2.43 \textpm 2.65 & 63.71\\
\multicolumn{1}{p{3.5cm}||}{\centering \textbf{Features + D + OH}} & \textbf{0} & 2.65 \textpm 1.58 & 69.31 & \textbf{0} & 3.16 \textpm 2.21 & 92.02 & 1.3 & 3.62 +- 2.45 & 62.08\\ 
\multicolumn{1}{p{3.5cm}||}{\centering \textbf{Features + E + OH}}& \textbf{0} & 2.62 \textpm 1.52 & 78.84 & \textbf{0} & 3.47 \textpm 2.28 & 89.15 & 27.6 & 4.2 \textpm 3.02 & 56.00\\ 		
\multicolumn{1}{p{3.5cm}||}{\centering \textbf{Features + D + E + OH}} & \textbf{0} & 2.31 \textpm 1.44 &  \textbf{96.65} &  \textbf{0} & 4.67 \textpm 2.91 & 91.05 & \textbf{2.6} & 4.37 \textpm 3.53 & 64.4\\ 
 \hline
\end{tabular}}
\caption{Ablation study results. D, E, and OH represents degree, entropy, and one hot encoded labels respectively. We vary node features along with different heuristic features to measure the effect of each of these features. Our proposed method \name is superior when it uses all the features.} 
\label{table:ablation_long}
\end{table*}

\subsection{Ablation Study}
\label{app:ablation}
  To infuse more information about the local graph structure and and its statistics, we use several heuristic features such as degree, entropy, and one-hot encoded labels (Refer \ref{sec:training}). We conduct an ablation study to investigate the effectiveness of each heuristic feature. Table \ref{table:ablation_long} summarises the findings. Our method is most consistent when it uses all features. Note that \textit{features} and \textit{entropy} together produce competitive results. However, the fidelity in BA-Shapes becomes much worse from this combination. This means, in most of the cases, this combination is unable to find the counterfactual example. In such cases, the possibility of getting better values in other measures increases.\\
\textbf{\textsc{Gcn} Vs. \gat:} \name uses a \gat to train the RL policy. In the next experiment, we evaluate the impact of replacing the \gat with a Graph Convolutional Network (\textsc{Gcn}). 
The results are presented in Table~\ref{table:policy_variation}. We see that \gat significantly outperforms \textsc{Gcn} in the inductive version and thereby justifying our choice.

\textbf{Heuristic Features:} Table \ref{table:ablation_long} contains an exhaustive analyses of the performance of \name-\textsc{inductive} using all combinations of heuristic features mentioned in section \ref{sec:induce}. The combination of features and entropy seems to allow best performance of the model on Tree-grid and Tree-cycles datasets, however as we see in Table \ref{table:dataset} that BA-shapes is a dense network and clearly the degree heuristic in combination with node features leads to excellent performance for BA-shapes in terms of the size and the accuracy. Since the ability of the method to find a counterfactual weighs more, our default model containing all heuristic combined with node features gives best overall performance in fidelity, with size and accuracy being better or comparable to the other combinations in most cases.

\noindent

\subsection{Additional Results on Counterfactual Size Distributions}
\label{app:counterfactual_size_disn} 
\begin{figure}[ht]
    \centering
    \subfloat[Inductive]{\includegraphics[width=1.6in]{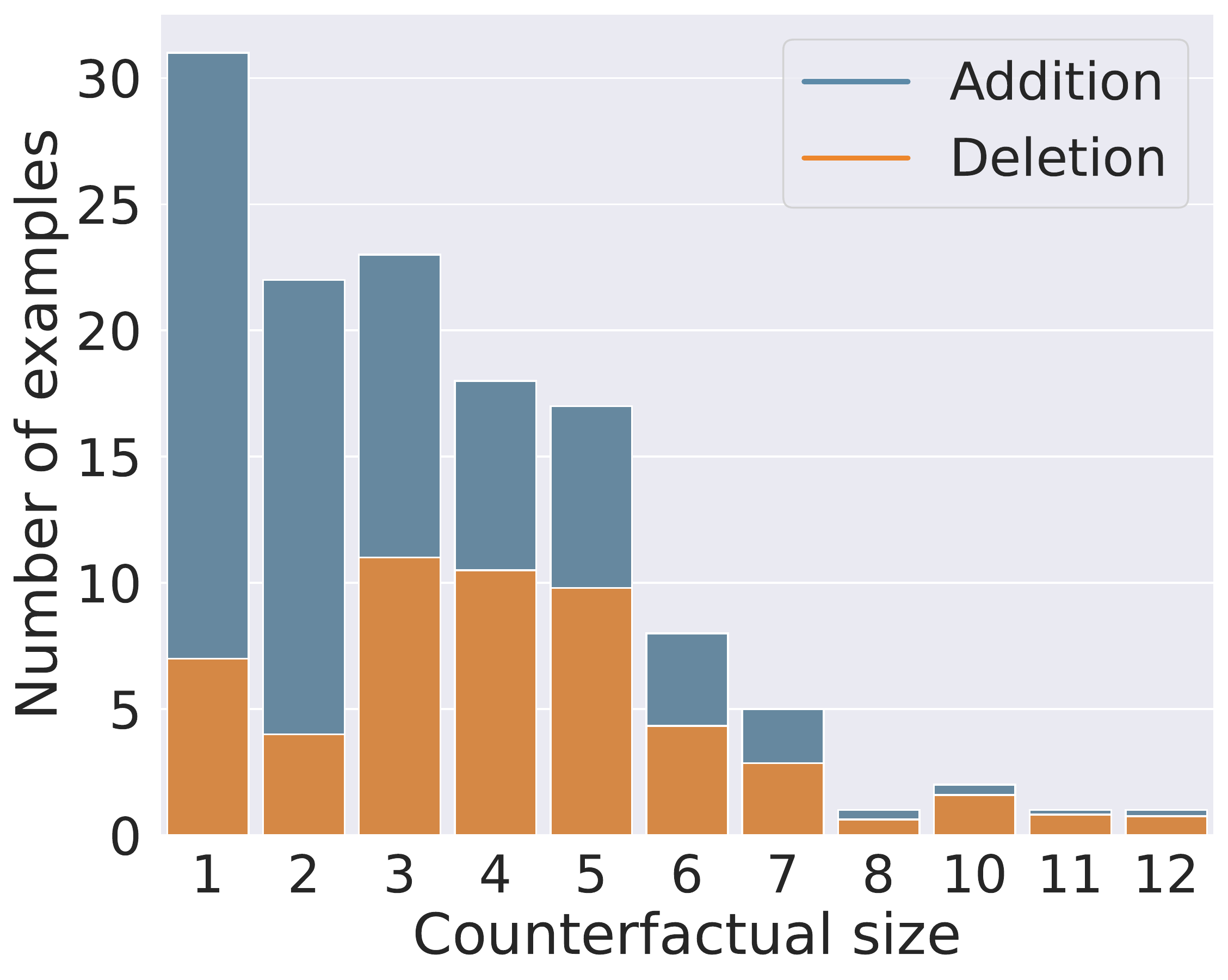}}
    \subfloat[Transductive]{\includegraphics[width=1.8in]{Plots/Syn4/transductive_non0_correct_only_bar_plot_train.pdf}}
    \caption{ The distributions of the edit size and their internal composition of edge additions and deletions by \name on the Tree-Grid dataset.}
\label{fig:tree_grid_ind_trans_cf_example}
\end{figure}

\begin{figure}[ht]
\vspace{-0.20in}
    \centering
    \subfloat[Inductive]{\includegraphics[width=1.6in]{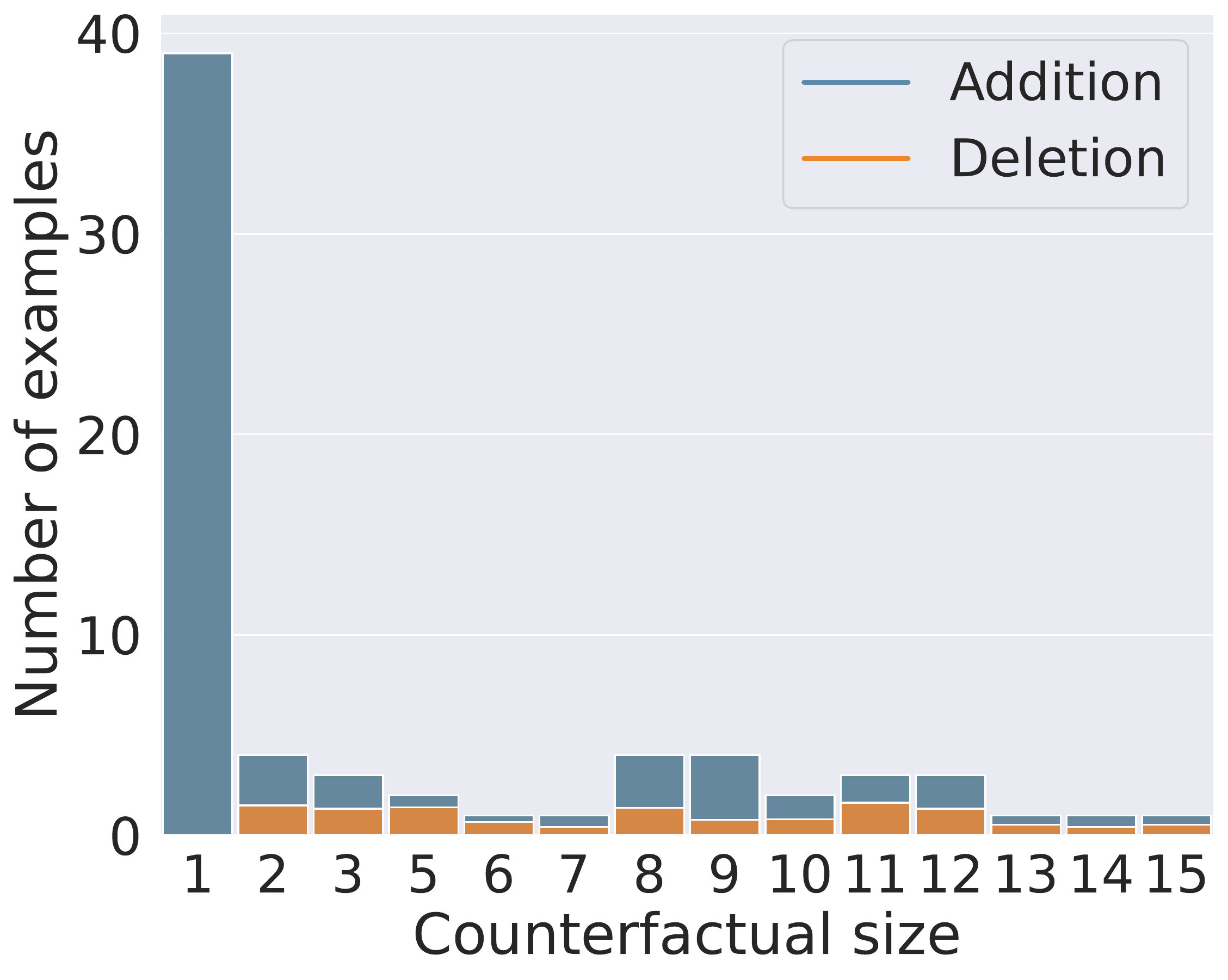}}
    \subfloat[Transductive]{\includegraphics[width=1.8in]{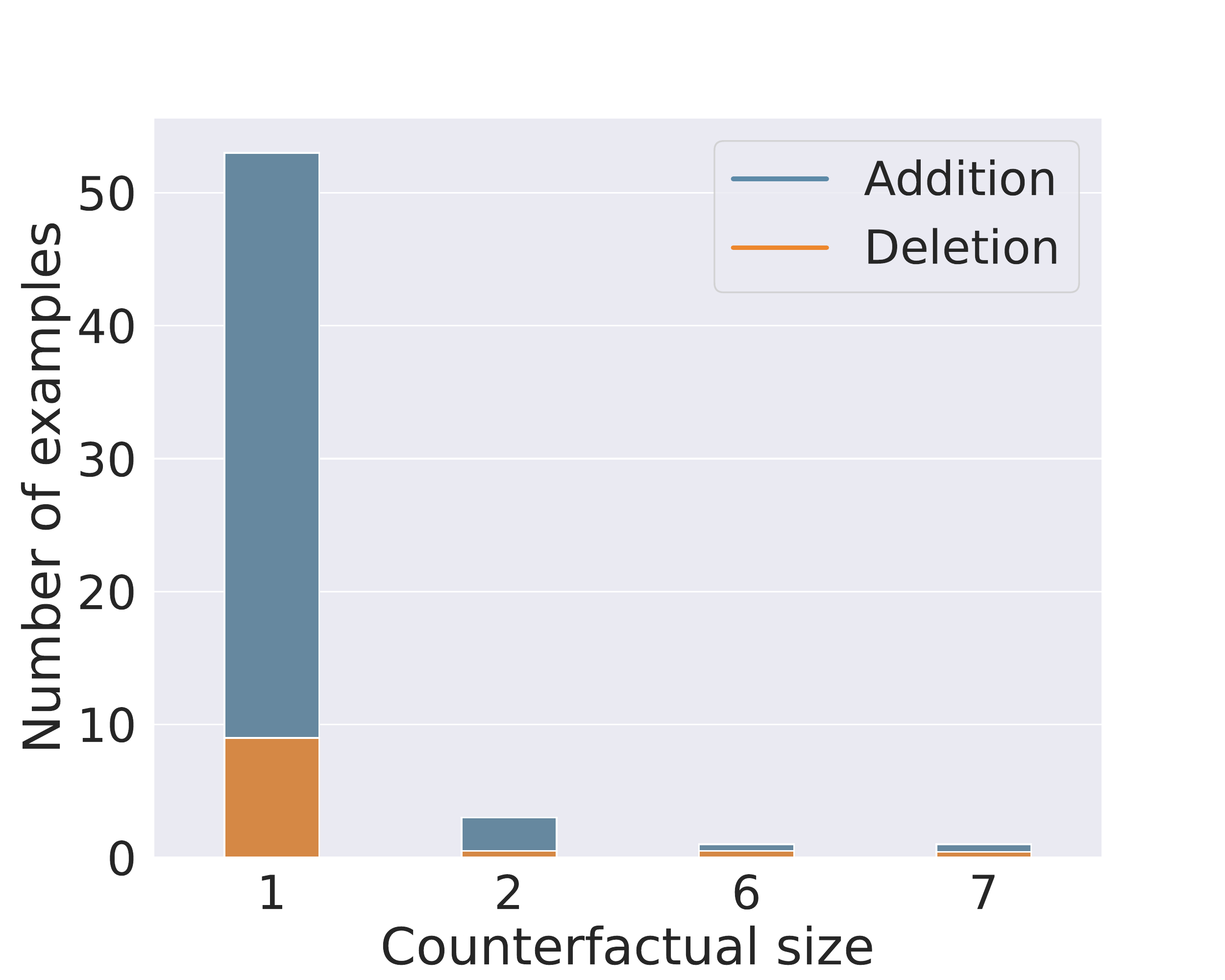}}
    \vspace{-0.10in}
    \caption{ The distributions of the edit size and their internal composition of edge additions and deletions by \name on the BA-Shapes dataset.}
\label{fig:ba_shapes_ind_trans_cf_example}
\end{figure}

In figures \ref{fig:tree_grid_ind_trans_cf_example} and \ref{fig:ba_shapes_ind_trans_cf_example} we observe the distributions of edit distance between the original and the counterfactual $h$-hop neighbourhood of instances in Tree-Grid and BA-Shapes datasets respectively. As observed both in inductive and transductive versions of \name most of the counterfactuals are of small size and dominated by edge additions. However, we can also observe that the transductive versions of \name does produce counterfactuals of size mostly localised around 1. This is because the parameters are tailored instance by instance. \name-inductive however with a minor trade-off in the counterfactual size, provides a comparable performance to \name-transductive (recall Tables \ref{table:node_class_results_transductive} and \ref{table:node_class_inductive}) while providing a speed-up of $79$x over all the transductive baselines (recall Table  \ref{table:running_time}). We further conduct experiments on how the accuracy of the explainer is affected with increasing counterfactual size in App. \ref{app:size_vs_acc}.

\sv{\subsection{Size vs. Accuracy Trade-off}
\label{app:size_vs_acc}
 The accuracy vs. counterfactual size trade-off for InduCE in Table \ref{table:size_vs_accuracy_ind} and \ref{table:size_vs_accuracy_tran}. We observe that with higher size, the accuracy decreases. Recall, we use benchmark datasets with ground-truth explanations where a node belongs to a particular class if it belongs to a certain motif. Hence, an explanation is accurate if it includes edges from the motif. We observe that when the explainer fails to find short explanations, it typically deviates towards a sequence of edges outside the motif. Hence the explainer fails to flip the label till a large set of edits are made.}

 \begin{table*}[h!]
\centering
\scalebox{1.0}{
\scriptsize
\begin{tabular}{ l||c||c||c } 
\toprule
\textbf{Size} & {\textbf{Acc. \% (Tree-Cycles)}}  & \textbf{Acc. \% (Tree-Grid)} & \textbf{Acc.\% (BA-Shapes)}\\ 
\midrule
1 & 100 & 100 & 100\\ 		
3 & 94 & 97 & 100\\ 
5 & 73 & 89 & 100\\ 
6 & 83 & NA & NA\\ 
7 & 86 & 88 & 100\\ 
10& NA & 80 & 100\\ 
15 & NA & 73 & 73\\ 
 \bottomrule
\end{tabular}}
\caption{\textbf{ Counterfactual Size vs. Accuracy Trade-off for \name - inductive: }The results suggest that as counterfactual size increases, the accuracy of the explanation decreases. NA stands for counterfactual of that size was not present.}
\label{table:size_vs_accuracy_ind}
\vspace{-0.10in}
\end{table*}

 \begin{table*}[h!]
\centering
\scalebox{1.0}{
\scriptsize
\begin{tabular}{ l||c||c||c } 
\toprule
 \textbf{Size} & {\textbf{Acc. \% (Tree-Cycles)}}  & \textbf{Acc. \% (Tree-Grid)} & \textbf{Acc. \% (BA-Shapes)}\\ 
\midrule
1 & 97 & 98 & 100\\ 		
2 & 100 & 100 & 100\\ 
6 & NA & NA & 50\\ 
7 & NA & NA & 57\\ 
 \bottomrule
\end{tabular}}
\caption{\textbf{ Counterfactual Size vs. Accuracy Trade-off for \name - transductive: }The results suggest that as counterfactual size increases, the accuracy of the explanation decreases. NA stands for counterfactual of that size was not present.}
\label{table:size_vs_accuracy_tran}
\vspace{-0.10in}
\end{table*}

\subsection{Baselines for ``ogbn-arxiv'' Dataset.}
\label{app:results_ogbn}
\newapp{\textbf{The baselines do not scale for Ogbn-arxiv dataset. We describe the details as follows. }Ogbn-arxiv is a \textbf{million-sized} node prediction dataset with \textbf{169,343} nodes and \textbf{1,166,243} edges. \cff and \gem do not scale (Recall Tables ~\ref{tab:transductive_ogbn} and ~\ref{tab:inductive_ogbn}) on this dataset since they employ computations on a dense adjacency matrix, which require $\mathcal{O}(n^2)$ space, where $n$ is the number of nodes in the graph. For a million-sized graph, this leads to memory overflow. Adapting to a sparse adjacency matrix requires non-trivial changes to the source code.\\}

\newapp{\cfg extracts the $k-$hop neighbourhood of a target node at runtime and adapts to sparse adjacency matrices more easily. \cfg's algorithm is model-agnostic, however, the code-base is suitable for its customized black-box and cannot be trivially extended to any other black-box. The explainer loads the black-box weights into itself before freezing them, assuming that the black-box uses the same architecutre as itself. It then uses those weights rather than the black-box \gnn during the explanation. In case the explainer's architecture does not match the black-box architecture, the keys for loading the weights do not match, hence, the explainer's weights are not loaded, rather randomly initialized. As a result, the explainer gets initialized with random weights rather than the black-box's weights and acts as a random classifier. This is an inefficient design choice and prevents \cfg's code from scaling for ogbn-arxiv(Recall Table ~\ref{tab:transductive_ogbn}). We use PyTorch-geometric's~\cite{fey2019fast} standard \textit{GCNConv} layers~\cite{kipf2016semi} that are compatible with sparse adjacency matrices to scale the black-box \gnn to the million-sized graph. \name's code is written in a model-agnostic fashion. Any black-box \gnn which employs \textit{log \softmax} non-linearity at the last layer is compatible with \name.}

\end{document}